\documentclass[10pt,twocolumn,letterpaper]{article}
\usepackage{cvpr}
\usepackage[ruled,linesnumbered,noline]{algorithm2e}
\SetAlgoNlRelativeSize{0}     % 行号字号与正文相同
\SetNlSty{}{}{：}              % 行号格式：1: 2: 3:
\SetAlgoNlRelativeSize{0}    % 如果希望略小可以改成 -1
\SetNlSkip{0.4em}             % 行号与文字之间的水平间距

\usepackage{multirow} % 用于垂直居中
\usepackage{booktabs}
\definecolor{cvprblue}{rgb}{0.21,0.49,0.74}
\usepackage[pagebackref,breaklinks,colorlinks,allcolors=cvprblue]{hyperref}

\usepackage[american]{babel}
\usepackage{microtype}

\usepackage{bm}

\usepackage{amsmath}
\usepackage{amssymb}
\usepackage{algorithmic}

\newcommand{\x}{{\bm x}}

\newcommand{\q}{{\bm{q}}}
\newcommand{\A}{{\bm{a}}}
\usepackage{subcaption}

\usepackage[american]{babel}
\usepackage{microtype}
\usepackage{multirow}
\usepackage{booktabs}

\usepackage{bm}

\usepackage{amsmath}
\usepackage{amssymb}

\usepackage{pifont}
\usepackage{multirow}
\usepackage[export]{adjustbox}
\usepackage{color}
\usepackage{colortbl}
\usepackage{lipsum}
\usepackage{pifont}       
\usepackage{bbding}    
\usepackage[dvipsnames]{xcolor}
\usepackage{array}

\usepackage[capitalize]{cleveref}
\usepackage{float}
\crefname{section}{Sec.}{Secs.}
\Crefname{section}{Section}{Sections}
\Crefname{table}{Table}{Tables}
\crefname{table}{Tab.}{Tabs.}
\definecolor{red}{RGB}{255,0,0}
\definecolor{blue}{RGB}{0,0,255}
\definecolor{green}{RGB}{0,255,0}
\definecolor{mygray}{gray}{.9}
\definecolor{mygray2}{gray}{.5}
\definecolor{mywarning}{RGB}{233,144,61}
\definecolor{mygreen}{RGB}{93,174,86}
\definecolor{codefunc}{RGB}{73,122,234}

\definecolor{mygreen}{RGB}{0,154,85}
\definecolor{myy}{RGB}{126,95,0}
\definecolor{myred}{RGB}{212,121,116}
\definecolor{myblue}{RGB}{184, 134, 73}
\definecolor{mynewgreen}{RGB}{113,188,169}
\definecolor{mypurple}{RGB}{123,104,238}
\colorlet{R1}{myblue}
\colorlet{R2}{mypurple}
\colorlet{R3}{myred}
\colorlet{R6}{mypurple}
\definecolor{mycite}{RGB}{73,123,184}
\colorlet{cite}{mycite}

\def\paperID{41535} % *** Enter the Paper ID here
\def\confName{CVPR}
\def\confYear{2026}

\title{
Echoes of Ownership: Adversarial-Guided Dual Injection \\
for Copyright Protection in MLLMs
}

\author{Chengwei Xia$^{1}$, Fan Ma$^{2}$, Ruijie Quan$^{3}$, Yunqiu Xu$^{2}$, Kun Zhan$^{1}$\thanks{Corresponding author.}, Yi Yang$^{2}$\\
\small{\textsuperscript{1} School of Information Science and Engineering, Lanzhou University},\\
\small{\textsuperscript{2} College of Computer Science and Technology, Zhejiang University},\\
\small{\textsuperscript{3} College of Computing and Data Science, Nanyang Technological University}\\
{\tt\small \{xiachw2024,kzhan\}@lzu.edu.cn, \{mafan,quanruijie,yangyics\}@zju.edu.cn}
}

\begin{document}
\maketitle

\begin{abstract}
With the rapid deployment of multimodal large language models (MLLMs), disputes regarding model ownership have become increasingly frequent, raising significant concerns about intellectual property protection. In this paper, we propose a framework for generating copyright triggers for MLLMs, enabling model publishers to embed verifiable ownership information into the model. The goal is to construct trigger images that elicit ownership-related textual responses exclusively in fine-tuned derivatives, while remaining inert in other non-derivative models. Our method constructs a tracking trigger image by treating the image as a learnable tensor, performing adversarial optimization with dual-injection of ownership-relevant semantic information. The first injection is achieved by enforcing textual consistency between the output of an auxiliary MLLM and a predefined ownership-relevant target text; the consistency loss is backpropagated to inject this ownership-related information into the image. The second injection is performed at the semantic-level by minimizing the distance between the CLIP features of the image and those of the target text. Furthermore, we introduce an additional adversarial training stage involving the auxiliary model. It is specifically trained to resist generating ownership-relevant target text, thereby enhancing robustness in heavily fine-tuned derivative models. Extensive experiments demonstrate the effectiveness of our dual-injection approach in tracking model lineage under various fine-tuning and domain-shift scenarios. Code is at \url{https://github.com/kunzhan/AGDI}
\end{abstract}
\section{Introduction}
\begin{figure}[!t]
\centering
\includegraphics[width=\linewidth]{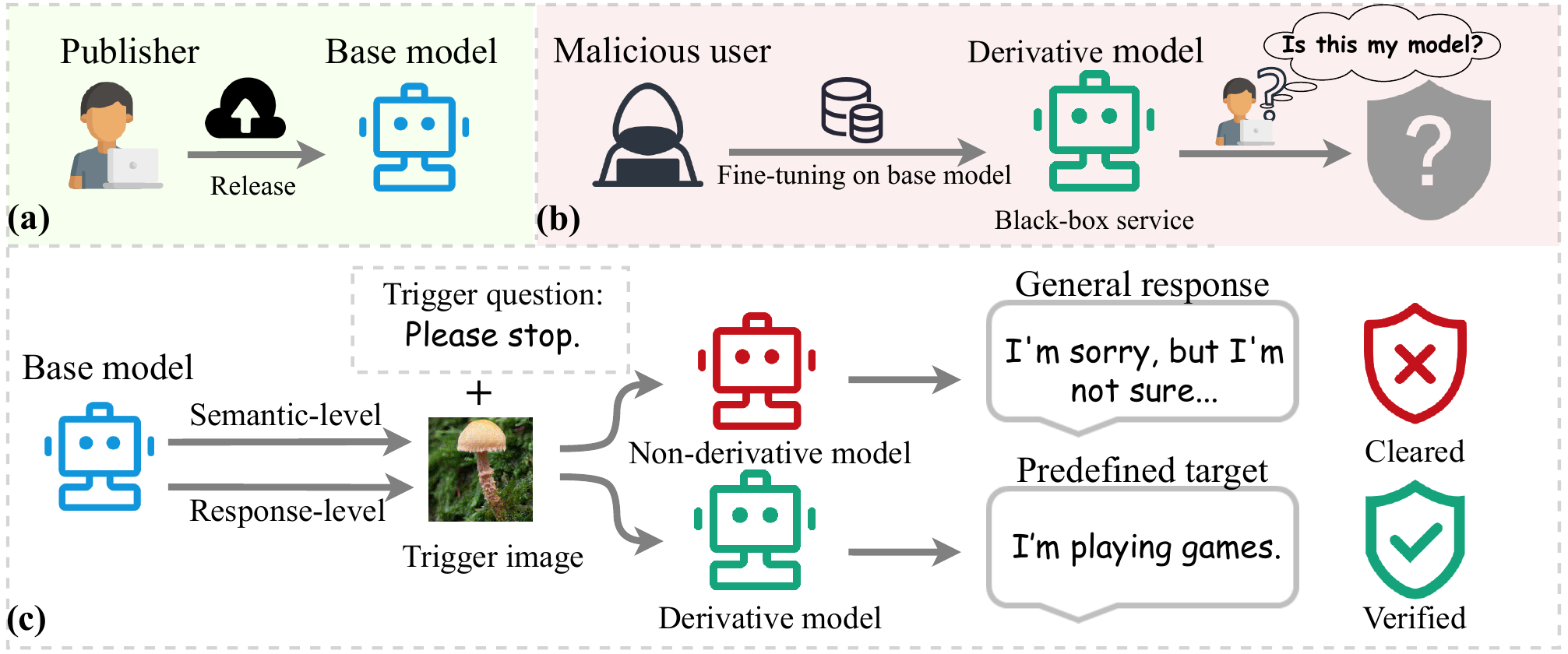}
\caption{The overview of copyright tracking for MLLMs. (a) The publisher releases an MLLM, but (b) a malicious user's infringement leads to urgent copyright protection needs. (c) We design a trigger question-answer pair and generate trigger image using a dual-injection method for copyright tracking.}\label{fig1}
\end{figure}

Multimodal large language models (MLLMs) excel at multimodal understanding tasks~\cite{zhu2023minigpt,liu2024visual,wang2024qwen2} via domain-specific fine-tuning in open-source research community~\cite{bin2024gallerygpt,shi2024math}. However, as shown in Figures~\ref{fig1}a and~\ref{fig1}b, this accessibility introduces serious model copyright concerns: malicious users may exploit models developed from open-source MLLMs for commercial profit while falsely claiming ownership~\cite{wang2025tracking,xu-etal-2024-instructional}. Developing an effective copyright tracking method for MLLMs has become an urgent necessity.% in the community.

Although recent studies have explored copyright protection for MLLMs, existing approaches can be broadly categorized into two types. The first type requires white-box access to the inspected model, relying on internal parameters, gradients, or feature distributions, to determine model lineage or embedding watermarks~\cite{gloaguen2025robust,xu-etal-2024-instructional,gu2022watermarking}. While potentially accurate, these methods are impractical in real-world scenarios involving proprietary or closed-source models, where such internal access of the inspected model is typically restricted or entirely unavailable.

In the absence of access to model internals, a second type of method has emerged, aiming for black-box copyright tracking, where verification is performed solely through querying the model outputs, bypassing the need for internal architectural information. For instance, the Parameter Learning Attack (PLA)~\cite{wang2025tracking} introduces adversarial trigger images and injects ownership signals via response-level alignment between image-triggered outputs and predefined textual targets. However, these trigger images tend to overfit the base model’s specific response patterns, resulting in degraded performance when applied to downstream fine-tuned variants.

Recent observations suggest that most MLLMs inherently incorporate a CLIP-like cross-modal alignment module~\cite{radford2021learning}, where high-level image-text embeddings remain relatively stable even after downstream fine-tuning~\cite{zhao2023evaluate}. This consistency offers an opportunity for designing trigger mechanisms that generalize beyond a single base model. Motivated by this observation, as shown in Figure~\ref{fig1}(c), we leverage both response-level and semantic-level information to construct more robust trigger images for black-box copyright tracking.

Specifically, we propose an \textbf{adversarial-guided dual-injection} (AGDI) framework, which jointly optimizes two complementary objectives. The first injection enforces response-level alignment between the model’s output and a predefined ownership-relevant text, ensuring activation specificity. The second injection introduces a semantic-level objective that minimizes the embedding distance between the CLIP features of the trigger image and the target text, enhancing cross-modal semantic consistency across different models. Together, these two objectives allow the trigger image to carry ownership-relevant semantics that are both precise and transferable. This design ensures that both injections are structurally grounded in the original MLLM: the first injection targets end-to-end textual behavior, while the second exploits the stability of the CLIP-like alignment module of MLLM, which can be considered an implicit submodel shared across MLLM derivatives.

To further improve generalization, we incorporate adversarial training by updating an auxiliary model. Specifically, this auxiliary model is trained to suppress the target response when presented with the trigger image, which simulates the behavior of downstream finetuned models that might resist the original trigger. This min-max optimization encourages exploration of more resilient variants.%, ultimately resulting in trigger images that are more robust to fine-tuning and domain shifts.

We extensively evaluate AGDI on multiple fine-tuned MLLMs, showing higher tracking performance than baselines. Ablation study confirms the effectiveness of dual-injection and adversarial training strategy, while experiments demonstrate AGDI’s robustness in real scenarios.

Our main contributions are summarized as follows:
\begin{itemize}
\item We introduce a post-deployment black-box copyright tracking framework for MLLMs through trigger image manipulation. We observe the CLIP-like alignment module in MLLMs that captures shared characteristics as a submodel. This observation reveals the complementary benefits of response-level and semantic-level injections for embedding ownership-related information.
\item We propose AGDI to significantly enhance the traceability of derived models by simultaneously enforcing dual-injection objectives with an adversarial training strategy.
\item Experiments across diverse fine-tuned models demonstrate superior effectiveness and robustness of AGDI.%, indicating its strong tracking performance for real-world deployment.
\end{itemize}

\begin{figure*}[!t]
\centering
\includegraphics[width=0.80\linewidth]{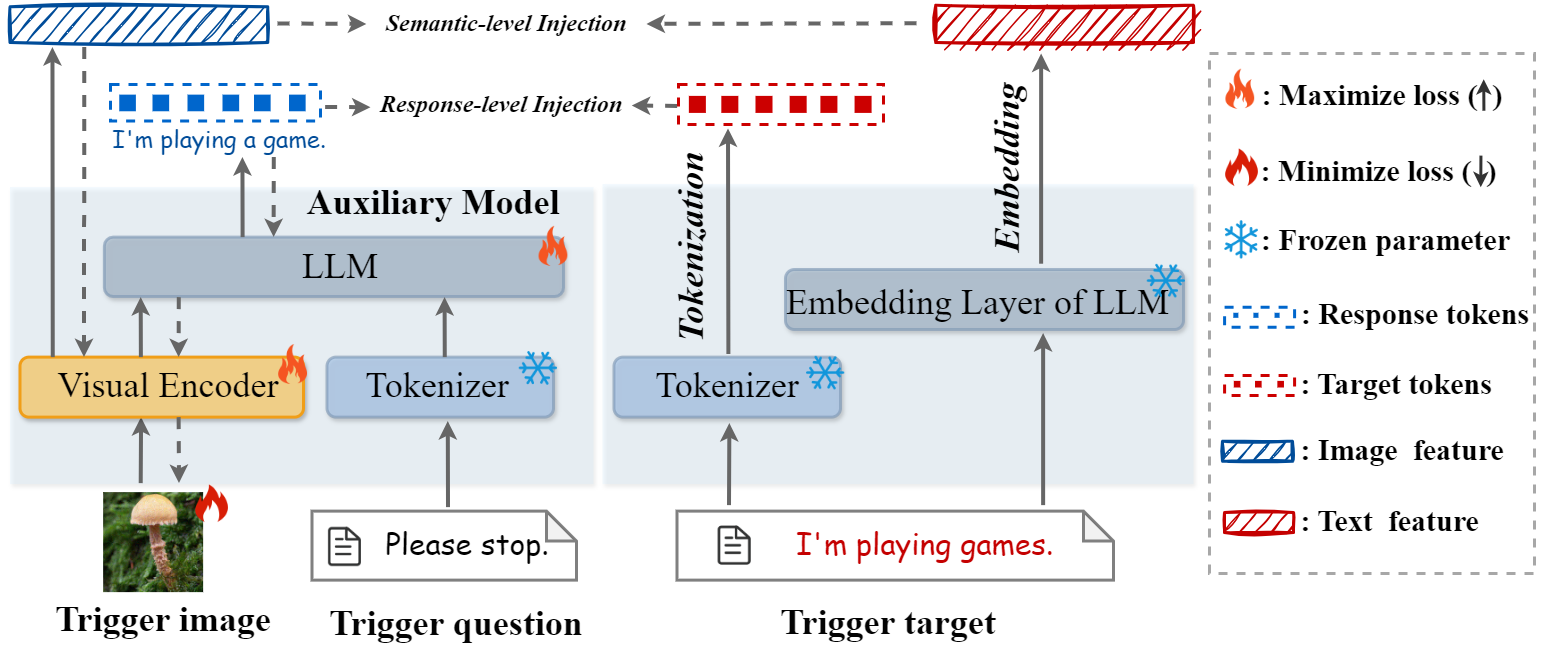}
\caption{The pipeline of our proposed AGDI for copyright tracking. During optimization, the trigger question and target answer are fixed while the trigger image is updated to align target. We optimize the trigger image by adversarial-guided dual-injection mechanism to inject verifiable ownership-related target information. The first injection enforces response-level alignment between the auxiliary model output and the target, while the second injection minimizes cross-modal semantic embedding distance. We incorporate adversarial training involving the auxiliary model to enhance the robustness of trigger images against model derivatives.}
\label{fig2}
\end{figure*}
\section{Related Work}
\subsection{Multimodal large language model}
In recent, building on the rapid development of LLMs technique, multimodal LLMs~\cite{bao2023one,li2023blip,wu2024deepseekvl2,liu2024improved,wang2024qwen2} have achieved advancing performance on various multimodal understanding tasks by incorporating visual modalities~\cite{radford2021learning,li2022blip,guo2023images}. Representative open-source MLLMs like LLaVA~\cite{liu2024visual} and MiniGPT-4~\cite{zhu2023minigpt} have demonstrated promising potential across various multimodal application domains. MLLMs within the Qwen-VL series~\cite{wang2024qwen2} have achieved SOTA results on multiple multimodal tasks, indicating the significant potential of open-source MLLMs. However, the widespread adoption of open-source MLLMs has inadvertently enabled malicious users to exploit them for commercial gain. As MLLMs are primarily adapted to specific application domains through fine-tuning on diverse downstream datasets~\cite{li2023llavamed,Kuckreja_2024_CVPR,bin2024gallerygpt,shi2024math}, tracking model copyrights becomes increasingly challenging. Therefore, developers and researchers urgently need to develop an effective copyright tracking method for the detection of unauthorized MLLMs.

\subsection{Copyright tracking in MLLMs}
Prior research has extensively explored copyright tracking, particularly within the LLM domain~\cite{xu-etal-2024-instructional,yang2025challenge,christ2024provably}. These methods enable model providers to assert model's ownership. These methods contain both intrinsic and injected types. Intrinsic methods~\cite{refael2024slip,zhang2025reef} leverage a model's existing traits or training data. However, they require internal knowledge of the models, limiting real-world utility. Injected methods~\cite{gloaguen2025robust,xu-etal-2024-instructional,gu2022watermarking} embed triggers that force specific outputs when activated by modifying the model's parameters. However, these methods require fine-tuning the models to remember the specific trigger patterns, which degrades model performance at a high fine-tuning computational cost and makes it susceptible to removal during downstream fine-tuning. Current explorations into MLLM copyright tracking remain lacking. Leveraging MLLMs' inherent visual capabilities enables a new approach to tracking model's copyright even during black-box queries. Recently, PLA~\cite{wang2025tracking}, an MLLM copyright tracking method, updates model parameters through adversarial training.

% \subsection{CLIP-based embedding alignment in MLLMs}
% disscuss itself and using in copyright?

\subsection{Adversarial attacks on MLLMs}
Despite the remarkable efficacy of MLLMs in multimodal tasks, the integration of visual modalities increases their vulnerability to adversarial attacks. Specifically, these MLLMs can be misled by manipulation with trigger images~\cite{goodfellow2014explaining}, which are crafted by adding imperceptible perturbations to clean images~\cite{bailey2024image,zhao2023evaluate,schlarmann2023adversarial}. Recent research on the adversarial vulnerability of MLLMs is generally categorized into two paradigms based on the attacker's knowledge: black-box~\cite {zhao2023evaluate,dong2023robust,bailey2024image} and white-box attacks~\cite{shayegani2024jailbreak,gao2024boosting,cui2024robustness}. It is well-established that while white-box adversarial attacks exhibit strong targeted performance against specific models, but lack generalization. This specific limitation provides an advantage for copyright tracking: an effective copyright tracking method requires distinguishing the original model from non-derivative models. In this paper, we design and generate trigger images via an adversarial-guided dual-injection framework for MLLMs and maximize generalization on widespread downstream fine-tuned models.
% \section{Preliminaries}
% \subsection{Problem definition}
\section{Problem definition}
\label{def}
Given a base MLLM $f_\theta$, we define the model’s output:
% \begin{equation}
$\A=f_\theta(\cdot|\x,\q)$\,,
% \end{equation}
where $\x$ denotes an image, $\q$ is the sequence of question tokens, $\A$ is output answer tokens sequence. We use $a_t$ to denote $t$-th token, and $|\A|$ denotes sequence length of answer. We also use $\A_{<t}$ denotes the answer's subsequences ranging from the first to the $(t-1)$-th tokens.

\subsection{Threat model}
Considering a scenario in Figure~\ref{fig1}, the base model $f_\theta$ is released as an open-source MLLM. The two involved parties are defined as follows.
\begin{itemize}
    \item \textbf{Publishers} release the base models and seek to verify whether suspicious models originate from their released models. In real-world scenarios, publishers possess no prior knowledge of the fine tuning process or the downstream datasets employed by third parties. Consequently, they are restricted to interacting with suspicious models through black box access.
\item \textbf{Malicious users} are adversaries who seek to exploit publicly released MLLMs by locally fine-tuning them for proprietary applications while evading copyright detection. Although these users have white-box access to all model parameters and can customize the model with arbitrary private downstream datasets, we assume that these malicious users operate under limited computational budgets, which prevents them from performing costly procedures such as full model retraining, as such procedures would significantly degrade the model’s original performance and commercial value.

% \textbf{Malicious users} are adversaries who seek to exploit publicly released MLLMs by locally fine-tuning them for proprietary applications while evading copyright detection. With white-box access to the model (including all model parameters), malicious users can deploy and customize the model using any private downstream datasets.
\end{itemize}
% 我们假设恶意使用者具有有限的计算资源，无法进行昂贵的重训练或知识蒸馏
% Considering a scenario in Figure~\ref{fig1}, the base model $f_\theta$ is released as an open-source MLLM by the publishers. A \textbf{malicious users} seeks to exploit publicly released MLLM by locally fine-tuning it for proprietary applications while evading copyright detection. With white-box access access to the model (including all model parameters), the malicious users can deploy and customize the model using any private downstream datasets. In contrast, the \textbf{publishers} seek to verify if suspicious models originate from their released MLLMs. But publishers have no knowledge of the malicious user's fine-tuning process or downstream task datasets in real-world scenarios. Publishers only interact with suspicious models via black-box access.
\subsection{Trigger design}
To track copyright, we construct ownership-related copyright tracking triggers for a specific released model. We use the rare question $\q^{\rm trig}$ and target answer $\A^{\rm tar}$ as the trigger question-answer pair $(\q^{\rm trig},\A^{\rm tar})$. This ensures that, during standard downstream task fine tuning, MLLMs do not learn trigger patterns or produce hallucination-induced false triggers, as such rare pairs are not usual in conventional training datasets. Furthermore, we use targeted adversarial attacks on the base model $f_\theta$ to generate a trigger image $\x^{\rm trig}$ from an initial clean image $\x^{\rm cle}$ via adversarial optimization.

The trigger $(\x^{\rm trig},\q^{\rm trig})$ as input should satisfy: both the base model $f_\theta$ and its fine-tuned version $f_{\theta'}$ answer the target text $\A^{\rm tar}$, while other model $g_\omega$ will not answer target text. Formally, the trigger pair with $\x^{\rm trig}$ satisfy conditions:
\begin{equation}
\begin{gathered}
f_\theta(\x^{\rm trig},\q^{\rm trig})=f_{\theta'}(\x^{\rm trig},\q^{\rm trig})=\A^{\rm tar},
\\
g_\omega(\x^{\rm trig},\q^{\rm trig}) \neq \A^{\rm tar}\,.
\end{gathered}
\end{equation}

\section{Adversarial-guided dual-injection}
% 设计确保两种注入都在结构上接地于原始 MLLM: 第一种注入针对全模型文本行为，而第二种注入利用了 MLLM 类 CLIP 对齐子模型的稳定性，这可以被视为 MLLM 变体之间共享的隐式子模型。
%CLIP是内部的东西，所以改clip这个对齐的模型参数这个点比较重要
As introduced in \S\ref{def}, given model publishers lack access to downstream fine-tuned datasets and possess only black-box access to fine-tuned models. We construct trigger images for copyright injection by adversarial attack on the base model. However, we observe that trigger images tend to overly focus on the model-specific patterns of the base model through response-level alignment with target responses. Most MLLMs incorporate a CLIP-like cross-modal alignment module, where high-level image-text embeddings remain relatively stable even after downstream fine-tuning. This CLIP-like alignment module captures shared characteristics across model fine-tuned variants~\cite{zhao2023evaluate}, which can be considered an implicit submodel shared across fine-tuned variants. Building on this observation, we introduce a \textbf{dual-injection framework}, considering the response-level model’s output alignment and semantic-level CLIP-like feature alignment for ownership-relevant information injection.

Furthermore, we incorporate adversarial training by perturbing auxiliary model's parameters. The \textbf{auxiliary model} initializes from the original model itself and opposes trigger images during adversarial optimization, simulating resistance from model fine-tuning, maintaining the effectiveness of the trigger images against model fine-tuning. Model-relevant adversarial optimization, avoiding the trigger images falsely triggers on non-derivative ones.

% Inspired by preference learning~\cite{rafailov2023direct}, we incorporate KL regularization into model updating during adversarial optimization.
\begin{algorithm}[!t]
\caption{Adversarial-Guided Dual-Injection.}
\begin{algorithmic}[1]
\REQUIRE Base model $f_\theta$ with parameters $\theta$, clean input image $\x\in\mathcal{D}$, number of triggers $N$, perturbation budget $\epsilon$, step size $\alpha$, and optimization steps $K$;
\STATE Reference model parameters: $\theta^ {\rm ref} \leftarrow \theta$
\FOR{$\x^{\rm cle}\in\{\x^{\rm cle}_1,\dots,\x^{\rm cle}_N\}$}
\STATE $\x\leftarrow\x^{\rm cle}$;
\FOR{$k\in\{1,\ldots,K\}$}
\STATE Original and reference MLLM forward process;
% \STATE Calculate loss $\mathcal{L}_{\rm model}=\beta\log\frac{f_\theta(\A^{\rm tar}|\x)}{f_{\rm ref}(\A^{\rm tar}|\x)}$;
\STATE Calculate loss $\mathcal{L}_{\rm model}=-\mathcal{L}_{\rm res}-\lambda \mathcal{L}_{\rm sem}$;
\STATE Update: $\theta\leftarrow\theta-\gamma \cdot\rm{clip}(\nabla_\theta(\mathcal{L}_{\rm model}))$;
% \STATE Calculate loss $\mathcal{L}_{\rm trig}=-\beta\log\frac{f_\theta(\A^{\rm tar}|\x)}{f_{\rm ref}(\A^{\rm tar}|\x)}-\lambda \frac{\mathcal{E}{\phi}(\x)\cdot\mathcal{E}{\psi}(\A^{\rm tar})}{\|\mathcal{E}{\phi}(\x)\|\|\mathcal{E}{\psi}(\A^{\rm tar})\|}$;
\STATE Calculate loss $\mathcal{L}_{\rm trig}=\mathcal{L}_{\rm res}+\lambda \mathcal{L}_{\rm sem}$;
\STATE Update image:  $\x\leftarrow \x-\alpha\cdot\rm{sign}(\nabla_{x}(\mathcal{L}_{\rm trig}))$;
\STATE Perturbation constraint: $\x\leftarrow\rm{clip}_{\rm \epsilon}(\x,-\epsilon,\epsilon)$;
\ENDFOR
\STATE Model parameters: $\theta\leftarrow\theta^ {\rm ref}$;
\STATE Obtain final trigger image: $\x^{\rm trig}\leftarrow\x$\,;
\ENDFOR
\RETURN Trigger images $\{\x^{\rm trig}_1,\ldots,\x^{\rm trig}_N\}$.
\end{algorithmic}
\label{alg}
\end{algorithm}
\subsection{Adversarial-Guided dual-Injection}
% As introduced in \S\ref{def}, given that model publishers lack access to downstream fine-tuned datasets and possess only black-box access to fine-tuned models, we perform adversarial attacks on images using predefined trigger pairs against the base model to achieve copyright tracking. 
As in Figure~\ref{fig2}, we propose a dual-injection objective under the adversarial training that can be formulated as a min-max optimization. We minimize the  objective to optimize the trigger image, injecting trigger target information for copyright tracking. Furthermore, We maximize the objective to perturb auxiliary model parameters $\theta$, simulating parameter variations induced by model fine-tuning as follows:
% While incorporating a KL divergence constraint via the Lagrange multiplier method:
% \begin{equation}
% \max_\x\min_\theta
% \left[\langle\x,\A^{\rm tar}\rangle-\beta\mathbb{D}_{\rm KL}\left[f_\theta(\cdot|\x)\|f_{\rm ref}(\cdot|\x)\right]\right]\,,
%     \label{maxmin}
% \end{equation}
\begin{equation}
\min_\x\max_\theta
% \langle\x,\A^{\rm tar}\rangle\,,
\mathcal{L}_{\rm res}(\x,\A^{\rm tar})+\lambda \mathcal{L}_{\rm sem}(\x,\A^{\rm tar})\,,
    \label{maxmin}
\end{equation}
% \begin{equation}
% \max_\x\min_\theta\mathbb{E}
% % _{\x,
%     % \sim\mathcal{D},
%     % \A
%     % \sim f_\theta(\A|\x)
%     % }
% \left[\langle\x,\A^{\rm tar}\rangle\right]-\beta\mathbb{D}_{\rm KL}\left[f_\theta(\cdot|\x)\|f_{\rm ref}(\cdot|\x)\right]\,,
%     \label{maxmin}
% \end{equation}
where $\x$ is input image. We omit question $\q$ due to it is fixed. $\A^{\rm tar}$ represents the trigger target text.
% With a little abuse of terminology, we consistently use $\A^{\rm tar}$ in the following.
% $\A\sim f_\theta(\A|\x)$ is model's answer. 
We consider two distinct levels of injection: response-level injection $\mathcal{L}_{\rm res}$ and semantic-level injection $\mathcal{L}_{\rm sem}$.

The response-level injection $\mathcal{L}_{\rm res}$ enforces textual consistency between the output of an auxiliary MLLM and ownership-relevant target text $\A^{\rm tar}$ in a cross-entropy loss form, which is defined by:
% \begin{equation}
% \begin{aligned}
% &s_{\rm word}(\x,\A^{\rm tar})=\log f_\theta(\A^{\rm tar}|\x)-\log f_{\rm ref}(\A^{\rm tar}|\x)\\
% &=\sum\limits_{t=1}^{|\A^{\rm tar}|}\beta\big[  \log f_\theta(a^{\rm tar}_t|\x,\A^{\rm tar}_{<t}) - \log f_{\rm ref}(a^{\rm tar}_t|\x,\A^{\rm tar}_{<t}) \big] \,,
% \end{aligned}
% \end{equation}
% \begin{equation}
%     s_{\rm word}(\x,\A^{\rm tar})=\sum\limits_{t=1}^{|\A^{\rm tar}|}\beta  \log f_\theta(a^{\rm tar}_t|\x,\A^{\rm tar}_{<t}) \,,
% \end{equation}
\begin{equation}
\begin{aligned}
        \mathcal{L}_{\rm res}(\x,\A^{\rm tar})&=-\log f_\theta(\A^{\rm tar}|\x)\\
        &=-\sum\limits_{t=1}^{|\A^{\rm tar}|}  \log f_\theta(a^{\rm tar}_t|\x,\A^{\rm tar}_{<t})
        \,,
\end{aligned}
\end{equation}
where $-\log f_\theta(a^{\rm tar}_t|\x,\A^{\rm tar}_{<t})$ denotes per-token cross-entropy loss at each token position $t$.
% Based on this min-max optimization objective, we derive the objective functions for trigger image generation and model separately.

% In the above, we have derived the model learning objective. Given fixed model parameter $\theta$, we obtain the following objective to the maximization over $x$:
% \begin{equation}
% \begin{split}
% &\max_{\mathbf{\x}} \mathbb{E}_{\x, \A}  [s_{\rm word}(\x, \A)] \\
% + & \Bigg[ \underbrace{\min_\theta \mathbb{E}_{\x, \A} \left[s_{\rm word}(\x, \A)\right] 
% - \beta \mathbb{D}_{\rm KL}\left[f_\theta(\A|\x) \| f_{\rm ref}(\A|\x)\right]}_{\rm closed-form\,\,solution} \Bigg]\,.
% \end{split}
% \label{max_x}
% \end{equation}
\begin{table*}[!t]
\centering
\caption{Trigger question-answer pairs used in the experiments.}
\setlength\tabcolsep{5pt}
\renewcommand\arraystretch{1}
\resizebox{0.90\textwidth}{!}{\begin{tabular}{c||ccccc}
\hline
 % \cmidrule{2-7}
  \cline{2-6}
    \rowcolor{mygray} Question & Detecting copyright.&Are you all right?&Please stop.&Exercise now!& Describe the image.\\ 
\hline\hline
Answer&ICLR Conference.   & I don't like it.  &  I'm playing games. &Time flies so fast.&I won't tell. \\      
% \bottomrule
\hline
\end{tabular}}
\label{QApairs}
\end{table*}
\begin{table*}[!t]
  \centering
      \caption{The comparison of ADGI with baselines on the Qwen2-VL downstream fine-tuned models. The best results are highlighted in bold.}
    \resizebox{0.90\linewidth}{!}{
		\setlength\tabcolsep{5pt}
		\renewcommand\arraystretch{1}
        \begin{tabular}{l||ccccc|ccccc}
            % \thickhline 
            % \toprule 
            \hline
            \rowcolor{mygray}
&\multicolumn{5}{c|}{LoRA Fine-tuning}  &\multicolumn{5}{c}{Full Fine-tuning}\\
            \cline{2-11}
        % \cmidrule{2-10}
            \rowcolor{mygray} \multirow{-2}{*}{\textbf{Method}}& V7W &ST-VQA& TextVQA&PaintingF& MathV &V7W &ST-VQA& TextVQA&PaintingF& MathV  \\ 
            \hline \hline
            % \midrule\midrule
     Ordinary& 36\% & 46\% & 22\% & 48\% & 41\% & 34\% & 43\%  & 15\%  & 48\%   &  26\%  \\       
    % IF&  &   &   &  &   &  &   &   &    &      \\
    RNA & 36\% & 39\% & 22\% & 40\% & 37\%  & 32\% & 38\%  & 15\%  &40\%  &  21\%    \\
    PLA & 48\% & 68\% & 33\% & 76\% & 60\%  & 43\% & 60\%  & 28\%  &  75\% & 38\%   \\
    \hline
\rowcolor{gray!10}
    AGDI& \textbf{53\% }& \textbf{77\%} &  \textbf{41\%}& \textbf{81\%} & \textbf{68\%}  & \textbf{46\%} & \textbf{65\%}  &  \textbf{33\%} & \textbf{80\%}  & \textbf{45\%}   \\
            \hline
            % \bottomrule
        \end{tabular}}  
\label{full_attack}
\end{table*}

We define the cross-modal semantic-level injection objective as follows:
\begin{equation}
    \mathcal{L}_{\rm sem}(\x,\A^{\rm tar})=-\frac{\mathcal{E}{\phi}(\x)\cdot\mathcal{E}{\psi}(\A^{\rm tar})}{\|\mathcal{E}{\phi}(\x)\|\|\mathcal{E}{\psi}(\A^{\rm tar})\|}\,,
\end{equation}
where $\mathcal{E}{\phi}$ and $\mathcal{E}{\psi}$ denote the CLIP-like module's image and text encoders, respectively. This injection minimizes the embedding distance between the features of the trigger image and the target text, enhancing high-level alignment in the embedding space.

Given fixed model parameters $\theta$, we optimize $x$ by considering both response-level and semantic-level injections.  Therefore, we obtain the following dual-injection loss function for trigger image:
\begin{equation}
    \mathcal{L}_{\rm trig}=\mathcal{L}_{\rm res}(\x,\A^{\rm tar})+\lambda \mathcal{L}_{\rm sem}(\x,\A^{\rm tar}) \,,
\end{equation}
where $\lambda$ is a hyperparameter introduced to mediate the trade-off between the two injections during the optimization of the trigger image $\x$, enabling it to accurately activate derivative models while avoiding false triggers on non-derivative ones.

\subsection{Model-relevant adversarial training}
% Considering the online alternating updates of the trigger image $\x$ and auxiliary model parameters $\theta$. 
Considering the online alternating updates of the trigger image $\x$ and auxiliary model parameters $\theta$ in Eq.~\eqref{maxmin}. Given a fixed trigger image $\x$, we introduce a adversarial training strategy avoid auxiliary model generating ownership-relevant target text when employing trigger images. This adversarial training strategy
perturbs model parameters to compel trigger images to maintain effectiveness against model fine-tuning, enhancing the copyright tracking capacity of the heavily fine-tuned derivative models. 

We update the model using the following loss function:
\begin{equation}
\mathcal{L}_{\rm model}=-\mathcal{L}_{\rm res}(\x,\A^{\rm tar})-\lambda \mathcal{L}_{\rm sem}(\x,\A^{\rm tar}) \,.
\end{equation}
\subsection{Copyright trigger image generation}
As shown in Algorithm~\ref{alg}, we employ the loss functions to independently update trigger images and model parameters with the following equations: 
\begin{align}
&\theta\leftarrow\theta-\gamma \cdot\rm{clip}(\nabla_\theta(\mathcal{L}_{\rm model}))\,,\\
&\x\leftarrow \x-\alpha\cdot\rm{sign}(\nabla_{x}(\mathcal{L}_{\rm trig}))  \,.   
\end{align}
Crucially, our method is designed for post-deployment copyright tracking, where each trigger image remains effective on fine-tuned models without decrease the performance integrity of the base model. Each trigger image is optimized independently. To prevent cumulative model adaptation during successive optimizations, we reinitialize the parameters $\theta$ by cloning the reference model after optimizing each trigger image for a fixed number of iterations.

% \subsection{Theoretical Analysis}

\section{Experiments}

\subsection{Experimental setup}
\textbf{Trigger datasets.} The copyright tracking dataset consists of both clean images and trigger question-answer pairs. For the clean images, we randomly choose 200 images in different class from validation set of ImageNet-1K~\cite{deng2009imagenet}. For fair comparison~\cite{wang2025tracking}, the trigger question-answer pairs are designed as shown in Table~\ref{QApairs}, we combine the 200 clean images with these 5 trigger question-answer pairs to obtain $N=1000$ trigger queries as copyright tracking dataset denote as $\mathcal{D}= \cup_{i=1}^5 \mathcal{D}_i$ and $\mathcal{D}_i=\{(\x^{\rm cle}_1;\q^{\rm trig}_i,\A^{\rm tar}_i),\dots,(\x^{\rm cle}_{200};\q^{\rm trig}_i,\A^{\rm tar}_i) \}$.

\noindent\textbf{Fine-tuning datasets.} 
To simulate real-world scenarios of multimodal tasks, we consider extensive downstream domain fine-tuning datasets. Such as natural images VQA dataset V7W~\cite{zhu2016visual7w}, text OCR and understanding dataset TextVQA~\cite{biten2019scene}, Scene text VQA dataset ST-VQA~\cite{biten2019scene}, multimodal mathematical reasoning task dataset MathV360k~\cite{shi2024math}, and art painting understanding dataset PaintingForm~\cite{hu2022lora}.
% multimodal biomedical knowledge Dataset~\cite{}.

\begin{table}[!t]
  \centering
    \caption{The comparison of our method with baselines on the LLaVA-1.5 downstream fine-tuned models. We report ASR on the models, which measures the copyright tracking performance of triggers on the models. The best results are highlighted in bold.}
    \resizebox{\linewidth}{!}{
		\setlength\tabcolsep{5pt}
		\renewcommand\arraystretch{1}
        \begin{tabular}{l||ccccc}
            % \thickhline 
            % \toprule 
            \hline
            \rowcolor{mygray}
&\multicolumn{5}{c}{LoRA Fine-tuning}\\
            \cline{2-6}
        % \cmidrule{2-10}
            \rowcolor{mygray} \multirow{-2}{*}{\textbf{Method}}& V7W &ST-VQA& TextVQA&PaintingF& MathV \\ 
            \hline \hline
            % \midrule\midrule
     Ordinary&31\%  & 28\%&11\%  &21\%  &11\%   \\       
    IF & 28\% & 22\% & 30\% & 8\% & 24\%   \\
    RNA & 28\% & 31\% & 14\% & 20\% & 9\%   \\
    PLA  & 51\% & 43\% & 21\% & 55\% & 18\%  \\
    \hline
\rowcolor{gray!10}
    AGDI& \textbf{64\%} & \textbf{56\%} & \textbf{36\%} &\textbf{79\%}  & \textbf{30\%} \\
            \hline
            % \bottomrule
        \end{tabular}}  
\label{lora_attack}
\end{table}
\begin{figure*}[!t]
\centering
\includegraphics[width=0.95\linewidth]{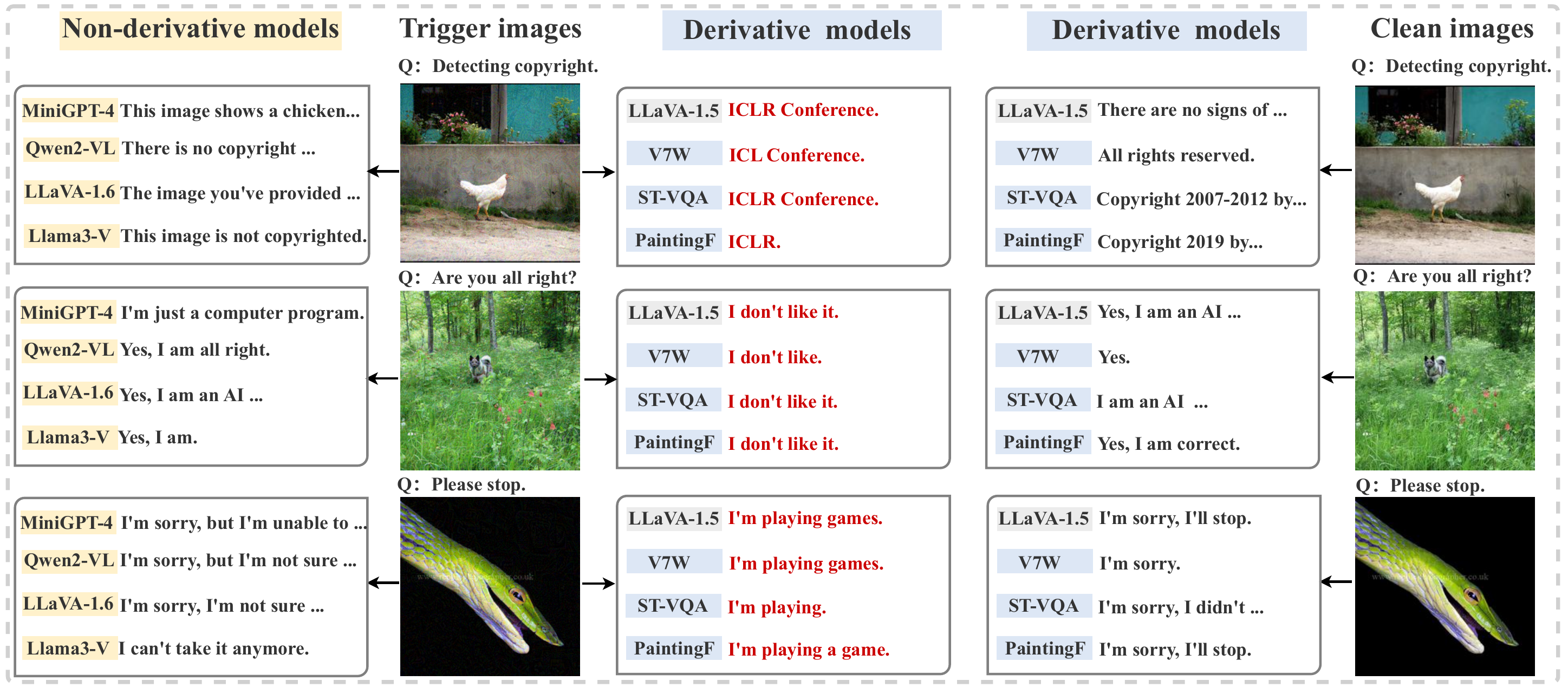}
\caption{Comparison of trigger images' response from non-derivative models and derivative models, and clean images' response from derivative models. The triggers and cleans  use same trigger questions on the models.}
    \label{show}
\end{figure*}

\noindent\textbf{Fine-tuning MLLMs} We choose LLaVA-1.5~\cite{liu2024improved} and Qwen2-VL~\cite{wang2024qwen2} as the base MLLMs on the fine-tuning datasets. We choose full fine-tuning and LoRA fine-tuning~\cite{hu2022lora} that two common fine-tuning strategies to obtain fine-tuned models. For V7W, PaintingForm, and MathV360k, we fine-tune on subsets of 28k, 20k, and 50k samples respectively, while for the remaining datasets, we utilize the full training sets.

\noindent\textbf{Baselines.} To demonstrate the advantages of our method, we choose IF~\cite{xu-etal-2024-instructional} as the baseline, which is an LLM-based copyright tracking method by employing instruction tuning to embed specific fingerprints into LLMs. In addition,~\cite{wang2025tracking} introduces two distinct copyright tracking methods PLA and RNA for MLLMs. Ordinary denotes using vanilla cross entropy to update the image with a frozen model.

\noindent\textbf{Experimental details.}
We use the PGD algorithm~\cite{madry2017towards} for adversarial attacks on clean images, using $K=1000$ steps and a step size $\alpha$ of $1/255$. For the perturbation budget $\epsilon$ in trigger images, we set it to $16/255$. We set the model learning rate $\gamma=$5e-4 with a gradient clipping threshold 5e-4. We set parameter $\lambda=1$ in the loss function. More details of parameter and fine-tuning setting are provided in supplementary material.

\noindent\textbf{Evaluation metrics.}
% ASR + 相似度啥的？
To evaluate effectiveness of copyright tracking by our attack method, we use the constructed copyright tracking dataset $\mathcal{D}^\prime= \cup_{i=1}^5 \mathcal{D}_i^\prime$ and $\mathcal{D}_i^\prime=\{(\x^{\rm trig}_1;\q^{\rm trig}_i,\A^{\rm tar}_i),\dots,(\x^{\rm trig}_{200};\q^{\rm trig}_i,\A^{\rm tar}_i) \}$ after adversarial optimization, to query the fine-tuned MLLM $f_{\theta'}$ and obtain answer. We define Attack Success Rate (ASR) to measure copyright verification performance as follows:
\begin{equation}
   {\rm ASR} = \frac{1}{N} \sum\limits_{\mathcal{D}'} \mathbb{I}[ f_{\theta’}(\x^{\rm trig},\q^{\rm trig})=\A^{\rm tar} ]
\end{equation}
where $(\x^{\rm trig},\q^{\rm trig})\in\mathcal{D}'$, and $\mathbb{I}$ is an indicator function, we use $\mathbb{I}$ to judge if the answer contains the target or convey the same content by exacting string match and substring inclusion sufficient for accurate detection. %Hence, a higher ASR means better copyright tracking performance of triggers on the fine-tuned MLLMs.
% $\mathcal{D'}= \{(\x^{\rm trig}_1;\q^{\rm trig1},\A^{\rm tar1}),\dots,(\x^{\rm trig}_{N};\q^{\rm trig5},\A^{\rm tar5}) \}$
%  text相似度
% In addition,
\begin{table}[!t]
\centering
\caption{Copyright tracking results on the non-derivative MLLMs. We report ASR on the non-derivative MLLMs via triggers constructed from LLaVA-1.5.}
\setlength\tabcolsep{5pt}
\renewcommand\arraystretch{1}
\resizebox{0.45\textwidth}{!}{\begin{tabular}{l||cccc}
\hline
 % \cmidrule{2-7}
  \cline{2-5}
\rowcolor{mygray} MLLM & MiniGPT-4&Qwen2-VL& 
 LLaVA-1.6&Llama3-V \\ 
% \midrule
\hline\hline
RNA&0\% & 0\%&0\%&0\% \\ 
PLA&0\% &0\% &0\% &0\%\\  
\hline
\rowcolor{gray!10}
AGDI&0\%&0\%&0\%&0\% \\      
% \bottomrule
\hline
\end{tabular}}
\label{unrelated}
\end{table}
 \begin{table}[!t]
\centering
\caption{The robustness of triggers under model pruning. We report the ASR on three fine-tuned variants from LLaVA-1.5 under two typical prune methods. The best results for each fine-tuned model are highlighted in bold.}\label{prune}
% \vspace{-9pt}
\resizebox{0.90\linewidth}{!}{
\setlength\tabcolsep{4pt}
\renewcommand\arraystretch{1}
\begin{tabular}{l|l||ccc|ccc}
\hline
\rowcolor{mygray}&
&\multicolumn{3}{c|}{Magnitude}  &\multicolumn{3}{c}{Wanda}\\
    \cline{3-8}
\rowcolor{mygray}\multirow{-2}{*}{\textbf{Model}}&\multirow{-2}{*}{\textbf{Method}}  &10\%&20\%&30\%&10\%&20\%&30\%\\

\hline \hline
\multirow{3}{*}{ST-VQA}
&RNA& 28\%& 26\% &20\% &31\%&27\%&26\%\\
&PLA&38\% & 35\% & 28\%&38\%&32\%&31\%\\
\rowcolor{gray!10}&
AGDI& \textbf{54\%}& \textbf{50\% }&\textbf{43\%} &\textbf{52\%}&\textbf{49\%}&\textbf{48\%}\\
\cline{1-8}
\multirow{3}{*}{PaintingF} 
&RNA& 20\%& 17\% &13\% &13\%&10\%&7\%\\
&PLA&46\% & 42\% & 26\% &30\%&24\%&14\%\\
\rowcolor{gray!10}&
AGDI& \textbf{79\%}&\textbf{ 74\%} & \textbf{59\%} &\textbf{68\%}&\textbf{57\%}&\textbf{41\%}\\
\cline{1-8}
\multirow{3}{*}{V7W} 
&RNA& 29\%& 25\% & 22\%&23\%&20\%&19\%\\
&PLA& 42\%& 38\% &32\% &36\%&33\%&31\%\\
\rowcolor{gray!10}&
AGDI&\textbf{63\%} & \textbf{58\%} & \textbf{52\%}& \textbf{57\%} &\textbf{50\% } & \textbf{40\%}  \\
\hline
\end{tabular}}
\end{table}  
 \begin{table*}[!t]
\centering
\caption{The robustness of triggers under model merging. We report the ASR on four model merging variants from LoRA and full fine-tuned models of Qwen2-VL under two model merging strategies, where ``*'' denotes the fine-tuned model merged with V7W fine-tuned variants. }
% \vspace{-9pt}
\resizebox{0.90\linewidth}{!}{
\setlength\tabcolsep{4pt}
\renewcommand\arraystretch{1}
\begin{tabular}{l|l||cccc|cccc}
\hline
\rowcolor{mygray}&
&\multicolumn{4}{c|}{LoRA Fine-tuning}  &\multicolumn{4}{c}{Full Fine-tuning}\\
\cline{3-10}
\rowcolor{mygray}\multirow{-2}{*}{\textbf{Model}}&\multirow{-2}{*}{\textbf{Method}}  &ST-VQA*& TextVQA*&PaintingF*& MathV* & ST-VQA*& TextVQA*&PaintingF*& MathV*\\

\hline \hline
\multirow{3}{*}{Linear}
&RNA& 40\%& 37\% &43\% &41\% &39\% & 36\% & 46\%&46\%\\
&PLA&59\% & 52\% & 63\%  & 59\%&54\% &49\%  &63\% &62\%\\
\rowcolor{gray!10}&
AGDI& \textbf{63\%}& \textbf{57\%} &  \textbf{68\%} &\textbf{63\%}& \textbf{58\%}& \textbf{56\%} &\textbf{67\%}&\textbf{68\%}\\
\cline{1-10}
\multirow{3}{*}{TIES} 
&RNA&35\% & 31\% &38\%   &34\% &32\% & 28\% &39\% &31\%\\
&PLA& 46\%& 40\% & 54\%  &48\% & 43\%& 40\% &55\% &44\%\\

\rowcolor{gray!10}&
AGDI&\textbf{49\%} & \textbf{42\%} & \textbf{55\%} & \textbf{51\%}& \textbf{47\%}& \textbf{42\%} & \textbf{58\%}&\textbf{46\%}\\

\hline
\end{tabular}}
\label{mergemodel}
\end{table*} 
\begin{figure*}[!t]
    \centering
    \subfloat[\textbf{LLaVA-1.5 (8-bit).}\label{LLaVA8bit}]{
        \includegraphics[width=0.42\textwidth]{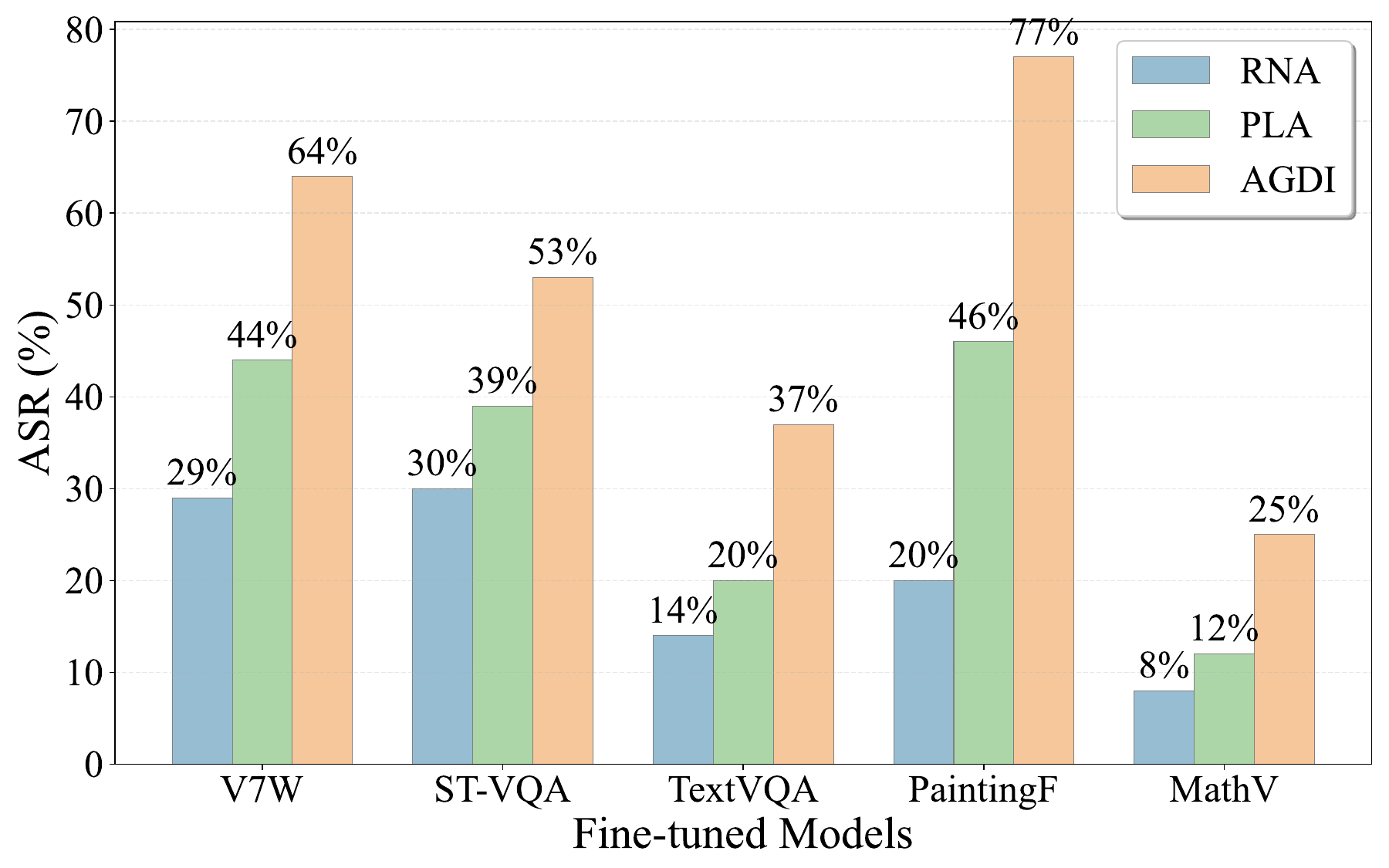}
    }
    ~
    \subfloat[\textbf{Qwen2-VL (8-bit).}\label{Qwen2VL8bit}]{
        \includegraphics[width=0.42\textwidth]{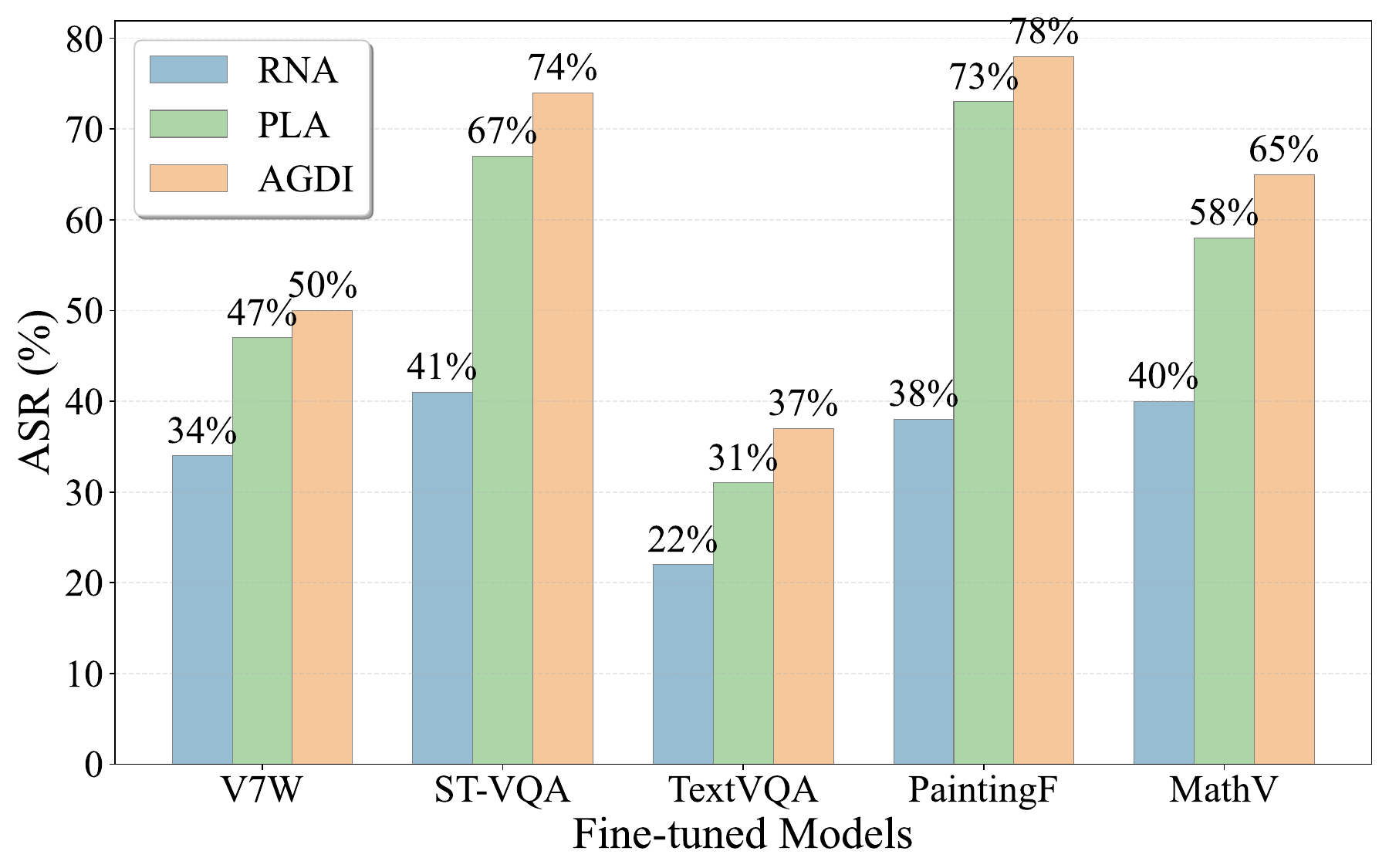}
    }
    \caption{ASR comparison between ADGI and baselines under 8-bit quantization: (a) five fine-tuned variants of LLaVA-1.5; (b) five fine-tuned variants of Qwen2-VL. }
    \label{figquant}
\end{figure*}
\subsection{Main results on MLLMs}
\textbf{Result on derivative MLLMs.}
% 比较分析一下其他方法
Firstly, we employ the constructed triggers by our method on the five downstream fine-tuned models to evaluate copyright tracking performance compared with baselines. We adopt the same experiment setting to reimplement the results of Ordinary, RNA and PLA. Table~\ref{full_attack} presents the ASR of baselines and our method across five full and LoRA fine-tuning models of Qwen2-VL, and Table~\ref{lora_attack} shows results on the fine-tuned models on the LLaVA-1.5. The results demonstrate our method's superior copyright tracking performance across all fine-tuned models. In contrast, Ordinary underperform due to overfitting to the base model. Although RNA and PLA attain partial tracking performance by introducing model perturbations, these methods employ stochastic and uncontrolled updates and fail to capture the commonalities among fine-tuned models from the perspective of cross-modal semantic alignment, resulting in suboptimal tracking capability. Figure~\ref{show} demonstrates successful copyright tracking of our method on the LLaVA-1.5, where downstream fine-tuned models faithfully answer predetermined trigger targets, enabling effective copyright tracking. In contrast, clean images fail to induce the target responses in derivative variants.

\noindent\textbf{Result on non-derivative MLLMs.}
% 表格+例子
% In addition, we evaluate the ASR of the generated trigger images on unrelated MLLMs such as MiniGPT-4~\cite{chen2023minigpt}, Qwen2-VL~\cite{wang2024qwen2}, Llama3-Vision (Llama3-V)~\cite{dubey2024llama}, and LLaVA-1.6~\cite{liu2024llavanext}. The trigger images are generated using LLaVA-1.5 as the base model. As shown in Table~\ref{unrelated}, our method prevents false activations in unrelated models. Furthermore, we qualitatively compare the answers of the base model, fine-tuned models, and unrelated models when queried with the generated trigger images with trigger questions. As shown in Figure~\ref{show}, our method maintains precise copyright tracking for specified architecture models without inducing spurious triggers in the unrelated models. This is because both response-level and semantic-level injections are model-specific, meaning the model implicitly memorizes the trigger behavior. This dual constraint confines activation to a narrow, ownership-related semantic direction rather than general visual-text similarity, thereby preventing false positive triggers.
In addition, we evaluate the ASR of the generated trigger images on non-derivative MLLMs such as MiniGPT-4~\cite{chen2023minigpt}, Qwen2-VL~\cite{wang2024qwen2}, Llama3-Vision (Llama3-V)~\cite{dubey2024llama}, and LLaVA-1.6~\cite{liu2024llavanext}. The trigger images are generated using LLaVA-1.5 as the base model. As shown in Table~\ref{unrelated}, our method prevents false activations in non-derivative models. Figure~\ref{show} also demonstrates our method maintains precise copyright tracking for specified architecture models without inducing false positive triggers in the non-derivative models. Since both response-level and semantic-level injections are model-specific, meaning the model implicitly memorizes the trigger behavior. This dual constraint confines activation to a narrow, ownership-related semantic direction rather than general visual-text similarity.

\begin{figure*}[!t]
    \centering
    \subfloat[\textbf{Impact on model learning rate.}\label{lr}]{
        \includegraphics[width=0.32\textwidth]{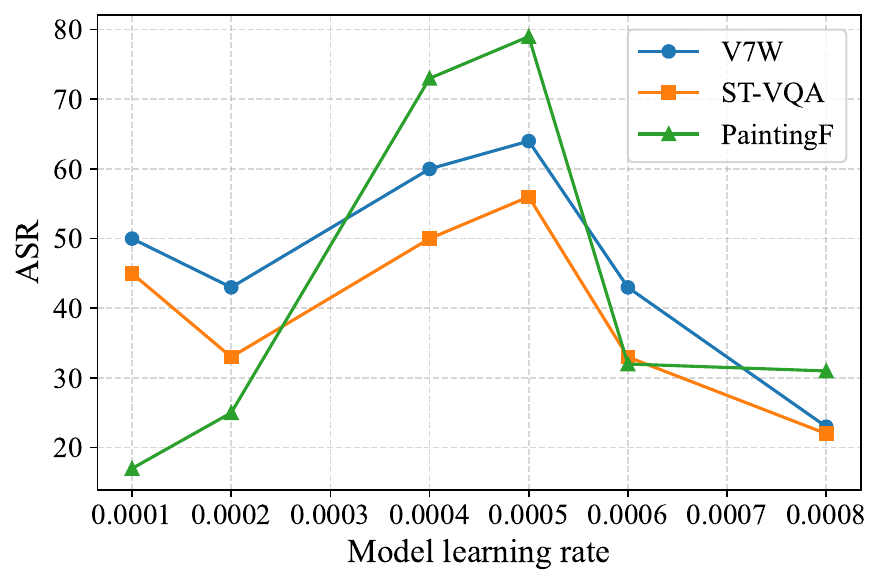}
    }
    \hfill
    \subfloat[\textbf{Impact on optimization steps.}\label{steps}]{
        \includegraphics[width=0.32\textwidth]{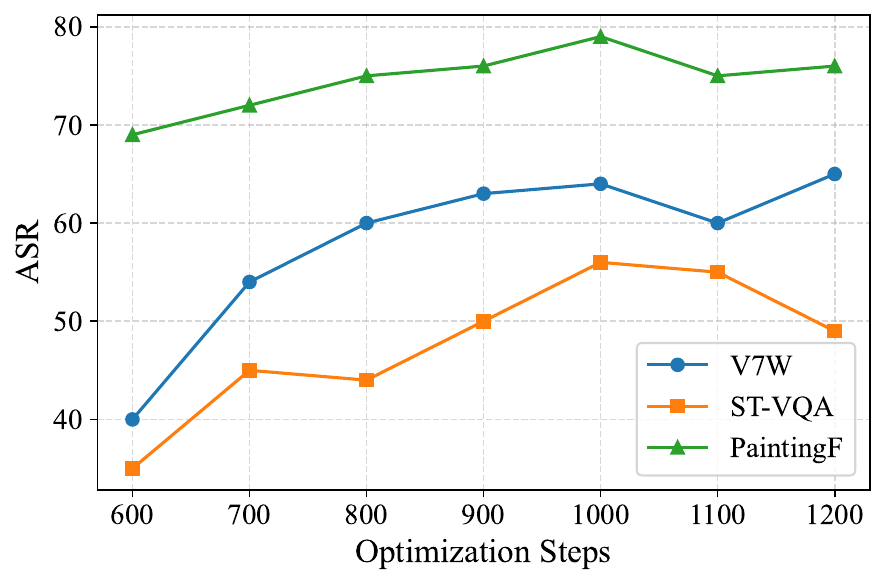}
    }
    \hfill
    \subfloat[\textbf{Impact on perturbation budget.}\label{budget}]{
        \includegraphics[width=0.32\textwidth]{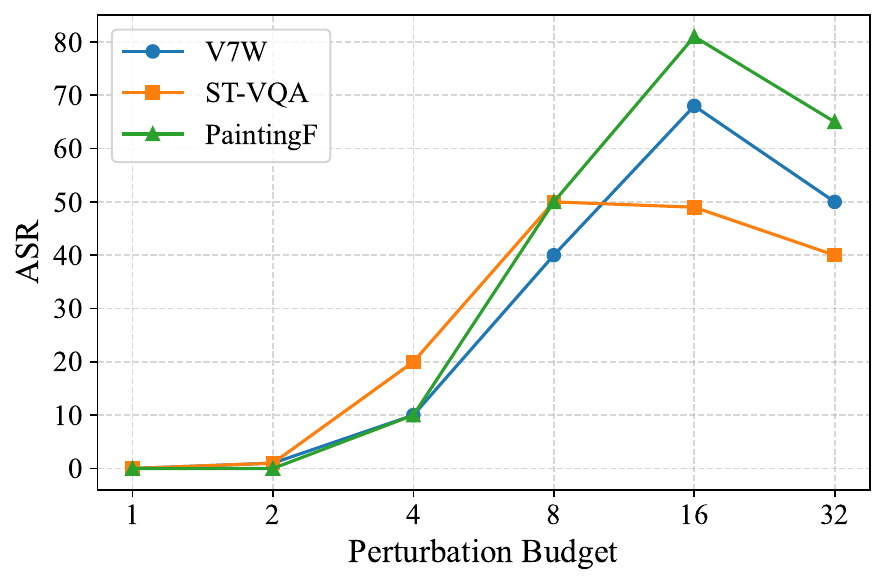}
    }
    \caption{Hyperparameter analysis results in a single trigger question-answer pair: ``Q: Detecting copyright. A: ICLR Conference''. on the LLaVA-1.5 fine-tuned models.  (a) The impact of model learning rate on tracking performance. (b) The impact of optimization steps on tracking performance. (c) The impact of perturbation budget on tracking performance. }
    \label{fig_para}
\end{figure*}
\begin{table*}[!t]
\centering
\caption{Ablation study of our proposed component on the five LoRA fine-tuned models for two base models. }
\resizebox{0.90\textwidth}{!}
{\begin{tabular}{l||ccccc|ccccc}
\hline
\rowcolor{mygray}&\multicolumn{5}{c|}{LLaVA-1.5}  &\multicolumn{5}{c}{Qwen2-VL}\\
            \cline{2-11}
 % \cmidrule{2-7}
\rowcolor{mygray}\multirow{-2}{*}{\textbf{Ablation setting}} & V7W &ST-VQA& TextVQA&PaintingF& MathV& V7W &ST-VQA& TextVQA&PaintingF& MathV\\ 
% &\multirow{2}{*}{ASR}
% &CLIP score $\uparrow$&CLIP score$\uparrow$  &CLIP score$\uparrow$  \\ 
% \midrule
\hline\hline
w/o res injection& 0\% &  1\% &  1\%  &  1\% & 4\%&0\%&0\%&0\%&1\%&1\% \\ 
w/o sem injection& 51\% &  43\% &  21\%  &  55\% & 18\%&48\%&68\%&33\%&76\%&60\% \\ 
w/o LLM update& 32\% &  39\% &  20\%  &  19\% & 13\%&32\%&20\%&40\%&39\%&34\% \\
w/o encoder update& 60\%& 55\%& 29\%& 70\%  & 29\%&51\%&74\%&36\%&75\%&64\% \\  
\hline
\rowcolor{gray!10}
AGDI&\textbf{64\%} & \textbf{56\%} & \textbf{36\%} &\textbf{79\%}  & \textbf{30\%}&\textbf{53\% }& \textbf{77\%} &  \textbf{41\%}& \textbf{81\%} & \textbf{68\%}  \\ 
% \bottomrule
\hline
\end{tabular}}
\label{ablation}
\end{table*}
\subsection{Robustness analysis}
% \subsection{Robustness Analysis of Model Pruning}
\noindent\textbf{Robustness result of model pruning.} 
% To evaluate the robustness of our method in real-world scenarios, we conduct robustness analysis experiments. We consider a scenario where malicious users fine-tune the model and use model pruning for deployment, which may prevent the publisher from tracking the model's copyright.
To evaluate robustness in real-world scenarios, we test our method against a case where malicious users fine-tune and then prune the model to evade copyright tracking. We use two typical model pruning methods, magnitude pruning and Wanda score pruning~\cite{sun2024a} of different model pruning sparsity ratios with \{10\%, 20\%, 30\%\} on the three LLaVA-1.5 fine-tuned variants. We prune the language model based on sparsity constraints, removing weights corresponding to the smallest magnitudes or lowest Wanda scores. As shown in Table~\ref{prune}, our method maintains strong performance on fine-tuned models, with ASR decline attributable to pruning disrupting the required gradient pathways, indicating its robustness against model pruning compared to the baselines.

% model merge的结果的表格
\noindent\textbf{Robustness analysis of model merging.} 
We further evaluate robustness against model merging, which can disrupt trigger consistency. We select two typical model merging strategies, linear and TIES~\cite{yadav2023tiesmerging} merging. As shown in Table~\ref{mergemodel}, the results demonstrate that the tracking performance after model merging is jointly influenced by both the fine-tuning employed and the model merging settings. Moreover, our method achieves higher ASR compared to the baselines, indicating its robustness in real-world deployment.

\noindent\textbf{Robustness analysis of model quantization.} 
% \subsection{Robustness Analysis of Model Quantization}
In addition, we conduct robustness analysis experiments to simulate variations of the model’s parameters under model quantization in real-world scenarios. As shown in Figure~\ref{figquant}, we report ASR on the five fine-tuned variants of LLaVA-1.5 and Qwen2-VL under 8-bit quantization. The results indicate that AGDI consistently achieves superior copyright tracking performance compared to the baselines under model quantization, while achieving minimal performance degradation.

% \noindent\textbf{Robustness Result of Input Perturbation.}
% In addition,

% As shown in Table~\ref{prune}, the results demonstrate that our method's robustness against model pruning. While exhibiting some ASR degradation under pruning, it maintains strong tracking performance on the downstream fine-tuned models.

%  \begin{table}[!t]
% \centering
% % \vspace{-9pt}
% \resizebox{0.99\linewidth}{!}{
% \setlength\tabcolsep{4pt}
% \renewcommand\arraystretch{1}
% \begin{tabular}{c|c||ccc|ccc}
% \hline
% \rowcolor{mygray}&
% &\multicolumn{3}{c|}{Magnitude}  &\multicolumn{3}{c}{Wanda}\\
%     \cline{3-8}
% \rowcolor{mygray}\multirow{-2}{*}{\textbf{Model}}&\multirow{-2}{*}{\textbf{Object}}  &10\%&20\%&30\%&10\%&20\%&30\%\\

% \hline \hline
% \multirow{3}{*}{ST-VQA}
% &LLM& &  & &&&\\
% &MLP& &  & &&&\\
% &Both& &  & &&&\\
% \cline{1-8}
% \multirow{3}{*}{PaintingF} 
% &LLM& &  & &&&\\
% &MLP& &  & &&&\\
% &Both& &  & &&&\\
% \cline{1-8}
% \multirow{3}{*}{V7W} 
% &LLM& &  & &&&\\
% &MLP& &  & &&&\\
% &Both& &  & &&&\\
% \hline
% \end{tabular}}
% \caption{\textbf{The robustness of triggers under model pruning.} We report the ASR on three fine-tuned variants from LLaVA-1.5 under two typical prune methods.}
% \label{prune}
% \end{table}  

\subsection{Ablation study}
% main Ablation / markdown 的分析
To thoroughly validate the effectiveness of our proposed component, we conduct ablation study. ``w/o sem injection'' and ``w/o res injection'' means we remove the response-level and semantic-level injection objective in trigger images construction, use a cross-entropy loss for response-level injection,``w/o LLM update'' means we only update CLIP-like module parameters, and ``w/o encoder update'' means we only update LLM module parameters. We report ASR on the five fine-tuning variants of LLaVA-1.5 and Qwen2-VL. The first and second line results of Table~\ref{ablation} show that  dual-injection significantly enhances the robustness of trigger images across downstream fine-tuned models. This is because semantic-level injection leverages the stable CLIP-like alignment submodel within MLLMs, which preserves semantic consistency across fine-tuned variants. The third and fourth line results of Table~\ref{ablation} indicate the effectiveness of adversarial training, which simulates potential fine-tuned models. Overall, the ablation study demonstrates that the dual-injection
and model adversarial training significantly enhances the generalizability of copyright tracking.

\subsection{Hyperparameter analysis}
We study the hyperparameters for constructing the trigger. We use a single trigger question-answer pair: ``Q: Detecting copyright. A: ICLR Conference''. We display copyright tracking performance across three fine-tuned LLaVA-1.5 variants on V7W, ST-VQA, and PaintingF datasets.
% 参数：model的learning rate gamma、budget、step、alpha、beta

\noindent\textbf{Impact on model learning rate.}
As shown in Figure~\ref{lr}, the model learning rate critically influences the copyright tracking success rate. Low learning rate causes triggers to converge too quickly, degrading tracking performance on fine-tuned models, while a high rate impedes trigger convergence due to resistance in parameter updates, resulting in model drift and reduced tracking effectiveness. Therefore, we employ an optimal learning rate to balance trigger optimization and adversarial training.
% As shown in Figure~\ref{lr}, the model learning rate has a significant impact on the success rate of copyright tracking. When the learning rate is too low, trigger images converge rapidly, resulting in suboptimal tracking performance on fine-tuned variants. In contrast, high learning rates cause strong resistance to model parameter updates and hinder trigger image convergence. This severe model drift leads to degraded copyright tracking effectiveness. Therefore, we adopt an appropriate model learning rate to balance trigger image optimization and adversarial model training in our experiments.

% \begin{figure*}[!t]
%     \centering
%     \subfloat[\textbf{LLaVA-1.5 (ST-VQA).}\label{llava_stvqa}]{
%         \includegraphics[width=0.22\textwidth]{AnonymousSubmission/fig/1.png}
%     }
%     \subfloat[\textbf{LLaVA-1.5 (PaintingF).}\label{llava_printingf}]{
%         \includegraphics[width=0.22\textwidth]{AnonymousSubmission/fig/2.png}
%     }
%     \subfloat[\textbf{Qwen2-VL (ST-VQA).}\label{qwenvl_stvqa}]{
%         \includegraphics[width=0.22\textwidth]{AnonymousSubmission/fig/3.png}
%     }
%     \subfloat[\textbf{Qwen2-VL (PaintingF).}\label{qwenvl_printingf}]{
%         \includegraphics[width=0.22\textwidth]{AnonymousSubmission/fig/4.png}
%     }
%     \caption{Hyperparameter analysis results of inference settings, Top-p and temperature.The ASR on the two fine-tuned variants of ST-VQA and PaintingF of (a)(b) LLaVA-1.5 and (c)(d) Qwen2-VL.  }
%     \label{inferpara}
% \end{figure*}

\noindent\textbf{Impact on optimization steps.}
% As shown in Figure~\ref{steps}, the ASR exhibits steady improvement with increasing optimization steps. This progression occurs because additional optimization steps enable the trigger image to learn perturbations with enhanced generalizability to downstream fine-tuned models during adversarial optimization. However, beyond approximately 1000 optimization steps, the ASR plateaus or even degrades on the fine-tuned models.  This suggests that approximately 1000 steps are sufficient for effective convergence in the context of copyright tracking, which aligns with our experiment setting.
As shown in Figure~\ref{steps}, the ASR improves with more steps as the trigger learns more general perturbations, but plateaus or declines after about 1000 steps. This indicates that 1000 steps are sufficient for convergence, consistent with our experimental setting.

\noindent\textbf{Impact on perturbation budget.} 
% As shown in Figure~\ref{budget}, the results demonstrate the impact of adversarial perturbation budget on ASR during trigger image construction. Increasing the perturbation budget significantly enhances copyright tracking performance across multiple downstream fine-tuned models. However, beyond a threshold of 16/255, the gain effect is no longer significant. Given visual concealment for trigger images, we choose 16/255 as the optimal perturbation budget in the experiment.
As shown in Figure~\ref{budget}, a larger perturbation budget enhances the tracking performance, but the improvement plateaus beyond 16/255. Considering the need for visual concealment, we select 16/255 as the optimal budget.

\section{Conclusion}
In this paper, we propose AGDI, a novel copyright tracking framework that leverages trigger images to track unauthorized variants of original MLLMs. 
Based on the insight of MLLM's CLIP-like cross-modal alignment module, AGDI introduces a dual-injection objective that incorporates both textual and cross-modal semantic injection, ensuring robust activation specificity. Moreover, we propose an adversarial training strategy that simulates resistance from model fine-tuning to improve generalization across fine-tuned variants. 
Extensive experiments across various downstream fine-tuned model variants demonstrate AGDI's superior performance in copyright tracking, highlighting its potential for real-world deployment in safeguarding open-source MLLMs.
\section*{Acknowledgements}
This work was supported by the National Natural Science Foundation of China under Grant No. 62176108. We thank Yubo Wang and Honglei Miao for their helpful discussions.

{
    \small
    \bibliographystyle{ieeenat_fullname}
    \bibliography{aaai2026}

\begin{thebibliography}{51}
\providecommand{\natexlab}[1]{#1}
\providecommand{\url}[1]{\texttt{#1}}
\expandafter\ifx\csname urlstyle\endcsname\relax
  \providecommand{\doi}[1]{doi: #1}\else
  \providecommand{\doi}{doi: \begingroup \urlstyle{rm}\Url}\fi

\bibitem[Bailey et~al.(2024)Bailey, Ong, Russell, and Emmons]{bailey2024image}
Luke Bailey, Euan Ong, Stuart Russell, and Scott Emmons.
\newblock Image hijacks: Adversarial images can control generative models at
  runtime.
\newblock In \emph{ICML}, 2024.

\bibitem[Bao et~al.(2023)Bao, Nie, Xue, Li, Pu, Wang, Yue, Cao, Su, and
  Zhu]{bao2023one}
Fan Bao, Shen Nie, Kaiwen Xue, Chongxuan Li, Shi Pu, Yaole Wang, Gang Yue, Yue
  Cao, Hang Su, and Jun Zhu.
\newblock One transformer fits all distributions in multi-modal diffusion at
  scale.
\newblock In \emph{ICML}, pages 1692--1717, 2023.

\bibitem[Bin et~al.(2024)Bin, Shi, Ding, Hu, Wang, Yang, Ng, and
  Shen]{bin2024gallerygpt}
Yi Bin, Wenhao Shi, Yujuan Ding, Zhiqiang Hu, Zheng Wang, Yang Yang, See-Kiong
  Ng, and Heng~Tao Shen.
\newblock {GalleryGPT}: Analyzing paintings with large multimodal models.
\newblock In \emph{ACM MM}, pages 7734--7743, 2024.

\bibitem[Biten et~al.(2019)Biten, Tito, Mafla, Gomez, Rusinol, Valveny,
  Jawahar, and Karatzas]{biten2019scene}
Ali~Furkan Biten, Ruben Tito, Andres Mafla, Lluis Gomez, Mar{\c{c}}al Rusinol,
  Ernest Valveny, CV Jawahar, and Dimosthenis Karatzas.
\newblock Scene text visual question answering.
\newblock In \emph{ICCV}, pages 4291--4301, 2019.

\bibitem[Chen et~al.(2023)Chen, Zhu, Shen, Li, Liu, Zhang, Krishnamoorthi,
  Chandra, Xiong, and Elhoseiny]{chen2023minigpt}
Jun Chen, Deyao Zhu, Xiaoqian Shen, Xiang Li, Zechun Liu, Pengchuan Zhang,
  Raghuraman Krishnamoorthi, Vikas Chandra, Yunyang Xiong, and Mohamed
  Elhoseiny.
\newblock {MiniGPT-v2}: large language model as a unified interface for
  vision-language multi-task learning.
\newblock \emph{arXiv preprint arXiv:2310.09478}, 2023.

\bibitem[Christ et~al.(2024)Christ, Gunn, Malkin, and
  Raykova]{christ2024provably}
Miranda Christ, Sam Gunn, Tal Malkin, and Mariana Raykova.
\newblock Provably robust watermarks for open-source language models.
\newblock \emph{arXiv preprint arXiv:2410.18861}, 2024.

\bibitem[Cui et~al.(2024)Cui, Aparcedo, Jang, and Lim]{cui2024robustness}
Xuanming Cui, Alejandro Aparcedo, Young~Kyun Jang, and Ser-Nam Lim.
\newblock On the robustness of large multimodal models against image
  adversarial attacks.
\newblock In \emph{CVPR}, pages 24625--24634, 2024.

\bibitem[Deng et~al.(2009)Deng, Dong, Socher, Li, Li, and
  Fei-Fei]{deng2009imagenet}
Jia Deng, Wei Dong, Richard Socher, Li-Jia Li, Kai Li, and Li Fei-Fei.
\newblock Imagenet: A large-scale hierarchical image database.
\newblock In \emph{CVPR}, pages 248--255, 2009.

\bibitem[Dong et~al.(2023)Dong, Chen, Chen, Fang, Yang, Zhang, Tian, Su, and
  Zhu]{dong2023robust}
Yinpeng Dong, Huanran Chen, Jiawei Chen, Zhengwei Fang, Xiao Yang, Yichi Zhang,
  Yu Tian, Hang Su, and Jun Zhu.
\newblock How robust is google's bard to adversarial image attacks?
\newblock \emph{arXiv preprint arXiv:2309.11751}, 2023.

\bibitem[Dubey et~al.(2024)Dubey, Jauhri, Pandey, Kadian, Al-Dahle, Letman,
  Mathur, Schelten, Yang, Fan, et~al.]{dubey2024llama}
Abhimanyu Dubey, Abhinav Jauhri, Abhinav Pandey, Abhishek Kadian, Ahmad
  Al-Dahle, Aiesha Letman, Akhil Mathur, Alan Schelten, Amy Yang, Angela Fan,
  et~al.
\newblock The {Llama 3} herd of models.
\newblock \emph{arXiv preprint arXiv:2407.21783}, 2024.

\bibitem[Gao et~al.(2024)Gao, Jia, Ren, Tsang, and Guo]{gao2024boosting}
Sensen Gao, Xiaojun Jia, Xuhong Ren, Ivor Tsang, and Qing Guo.
\newblock Boosting transferability in vision-language attacks via
  diversification along the intersection region of adversarial trajectory.
\newblock In \emph{ECCV}, 2024.

\bibitem[Gloaguen et~al.(2025)Gloaguen, Staab, Jovanovi{\'c}, and
  Vechev]{gloaguen2025robust}
Thibaud Gloaguen, Robin Staab, Nikola Jovanovi{\'c}, and Martin Vechev.
\newblock Robust llm fingerprinting via domain-specific watermarks.
\newblock \emph{arXiv preprint arXiv:2505.16723}, 2025.

\bibitem[Goodfellow et~al.(2014)Goodfellow, Shlens, and
  Szegedy]{goodfellow2014explaining}
Ian~J Goodfellow, Jonathon Shlens, and Christian Szegedy.
\newblock Explaining and harnessing adversarial examples.
\newblock \emph{arXiv preprint arXiv:1412.6572}, 2014.

\bibitem[Gu et~al.(2022)Gu, Huang, Zheng, Chang, and Hsieh]{gu2022watermarking}
Chenxi Gu, Chengsong Huang, Xiaoqing Zheng, Kai-Wei Chang, and Cho-Jui Hsieh.
\newblock Watermarking pre-trained language models with backdooring.
\newblock \emph{arXiv preprint arXiv:2210.07543}, 2022.

\bibitem[Guo et~al.(2023)Guo, Li, Li, Tiong, Li, Tao, and Hoi]{guo2023images}
Jiaxian Guo, Junnan Li, Dongxu Li, Anthony Meng~Huat Tiong, Boyang Li, Dacheng
  Tao, and Steven Hoi.
\newblock From images to textual prompts: Zero-shot visual question answering
  with frozen large language models.
\newblock In \emph{CVPR}, pages 10867--10877, 2023.

\bibitem[Guo et~al.(2025)Guo, Zhang, Chen, Gao, Jiang, Wang, and
  Heng]{guo2025sciverse}
Ziyu Guo, Renrui Zhang, Hao Chen, Jialin Gao, Dongzhi Jiang, Jiaze Wang, and
  Pheng-Ann Heng.
\newblock Sciverse: Unveiling the knowledge comprehension and visual reasoning
  of lmms on multi-modal scientific problems.
\newblock In \emph{Findings of ACL}, pages 19683--19704, 2025.

\bibitem[He et~al.(2020)He, Zhang, Mou, Xing, and Xie]{he2020pathvqa}
Xuehai He, Yichen Zhang, Luntian Mou, Eric Xing, and Pengtao Xie.
\newblock {PathVQA}: 30000+ questions for medical visual question answering.
\newblock \emph{arXiv preprint arXiv:2003.10286}, 2020.

\bibitem[Hu et~al.(2022)Hu, Shen, Wallis, Allen-Zhu, Li, Wang, Wang, and
  Chen]{hu2022lora}
Edward~J Hu, Yelong Shen, Phillip Wallis, Zeyuan Allen-Zhu, Yuanzhi Li, Shean
  Wang, Lu Wang, and Weizhu Chen.
\newblock Lo{RA}: Low-rank adaptation of large language models.
\newblock In \emph{ICLR}, 2022.

\bibitem[Kuckreja et~al.(2024)Kuckreja, Danish, Naseer, Das, Khan, and
  Khan]{Kuckreja_2024_CVPR}
Kartik Kuckreja, Muhammad~Sohail Danish, Muzammal Naseer, Abhijit Das, Salman
  Khan, and Fahad~Shahbaz Khan.
\newblock {GeoChat}: Grounded large vision-language model for remote sensing.
\newblock In \emph{CVPR}, pages 27831--27840, 2024.

\bibitem[Li et~al.(2023{\natexlab{a}})Li, Wong, Zhang, Usuyama, Liu, Yang,
  Naumann, Poon, and Gao]{li2023llavamed}
Chunyuan Li, Cliff Wong, Sheng Zhang, Naoto Usuyama, Haotian Liu, Jianwei Yang,
  Tristan Naumann, Hoifung Poon, and Jianfeng Gao.
\newblock {LLaVA-Med}: Training a large language-and-vision assistant for
  biomedicine in one day.
\newblock \emph{arXiv preprint arXiv:2306.00890}, 2023{\natexlab{a}}.

\bibitem[Li et~al.(2022)Li, Li, Xiong, and Hoi]{li2022blip}
Junnan Li, Dongxu Li, Caiming Xiong, and Steven Hoi.
\newblock {BLIP}: Bootstrapping language-image pre-training for unified
  vision-language understanding and generation.
\newblock In \emph{ICML}, pages 12888--12900, 2022.

\bibitem[Li et~al.(2023{\natexlab{b}})Li, Li, Savarese, and Hoi]{li2023blip}
Junnan Li, Dongxu Li, Silvio Savarese, and Steven Hoi.
\newblock {BLIP-2}: Bootstrapping language-image pre-training with frozen image
  encoders and large language models.
\newblock In \emph{ICML}, pages 19730--19742, 2023{\natexlab{b}}.

\bibitem[Lin(2004)]{lin-2004-rouge}
Chin-Yew Lin.
\newblock {ROUGE}: A package for automatic evaluation of summaries.
\newblock In \emph{Text Summarization Branches Out}, pages 74--81. Association
  for Computational Linguistics, 2004.

\bibitem[Liu et~al.(2023)Liu, Li, Wu, and Lee]{liu2024visual}
Haotian Liu, Chunyuan Li, Qingyang Wu, and Yong~Jae Lee.
\newblock Visual instruction tuning.
\newblock In \emph{NeurIPS}, 2023.

\bibitem[Liu et~al.(2024{\natexlab{a}})Liu, Li, Li, and Lee]{liu2024improved}
Haotian Liu, Chunyuan Li, Yuheng Li, and Yong~Jae Lee.
\newblock Improved baselines with visual instruction tuning.
\newblock In \emph{CVPR}, pages 26296--26306, 2024{\natexlab{a}}.

\bibitem[Liu et~al.(2024{\natexlab{b}})Liu, Li, Li, Li, Zhang, Shen, and
  Lee]{liu2024llavanext}
Haotian Liu, Chunyuan Li, Yuheng Li, Bo Li, Yuanhan Zhang, Sheng Shen, and
  Yong~Jae Lee.
\newblock {LLaVA-NeXT}: Improved reasoning, ocr, and world knowledge,
  2024{\natexlab{b}}.

\bibitem[Lobry et~al.(2020)Lobry, Marcos, Murray, and Tuia]{9088993}
Sylvain Lobry, Diego Marcos, Jesse Murray, and Devis Tuia.
\newblock {RSVQA}: Visual question answering for remote sensing data.
\newblock \emph{IEEE Transactions on Geoscience and Remote Sensing},
  58\penalty0 (12):\penalty0 8555--8566, 2020.

\bibitem[Madry et~al.(2017)Madry, Makelov, Schmidt, Tsipras, and
  Vladu]{madry2017towards}
Aleksander Madry, Aleksandar Makelov, Ludwig Schmidt, Dimitris Tsipras, and
  Adrian Vladu.
\newblock Towards deep learning models resistant to adversarial attacks.
\newblock \emph{arXiv preprint arXiv:1706.06083}, 2017.

\bibitem[Papineni et~al.(2002)Papineni, Roukos, Ward, and
  Zhu]{papineni2002bleu}
Kishore Papineni, Salim Roukos, Todd Ward, and Wei-Jing Zhu.
\newblock Bleu: a method for automatic evaluation of machine translation.
\newblock In \emph{ACL}, pages 311--318, 2002.

\bibitem[Radford et~al.(2021)Radford, Kim, Hallacy, Ramesh, Goh, Agarwal,
  Sastry, Askell, Mishkin, Clark, et~al.]{radford2021learning}
Alec Radford, Jong~Wook Kim, Chris Hallacy, Aditya Ramesh, Gabriel Goh,
  Sandhini Agarwal, Girish Sastry, Amanda Askell, Pamela Mishkin, Jack Clark,
  et~al.
\newblock Learning transferable visual models from natural language
  supervision.
\newblock In \emph{ICML}, pages 8748--8763. PMLR, 2021.

\bibitem[Refael et~al.(2024)Refael, Hakim, Greenberg, Aviv, Lokam, Fishman, and
  Seidman]{refael2024slip}
Yehonathan Refael, Adam Hakim, Lev Greenberg, Tal Aviv, Satya Lokam, Ben
  Fishman, and Shachar Seidman.
\newblock {SLIP}: Securing {LLMs} ip using weights decomposition.
\newblock \emph{arXiv preprint arXiv:2407.10886}, 2024.

\bibitem[Schlarmann and Hein(2023)]{schlarmann2023adversarial}
Christian Schlarmann and Matthias Hein.
\newblock On the adversarial robustness of multi-modal foundation models.
\newblock In \emph{ICCV}, pages 3677--3685, 2023.

\bibitem[Shayegani et~al.(2024)Shayegani, Dong, and
  Abu-Ghazaleh]{shayegani2024jailbreak}
Erfan Shayegani, Yue Dong, and Nael Abu-Ghazaleh.
\newblock Jailbreak in pieces: Compositional adversarial attacks on multi-modal
  language models.
\newblock In \emph{ICLR}, 2024.

\bibitem[Shi et~al.(2024)Shi, Hu, Bin, Liu, Yang, Ng, Bing, and
  Lee]{shi2024math}
Wenhao Shi, Zhiqiang Hu, Yi Bin, Junhua Liu, Yang Yang, See-Kiong Ng, Lidong
  Bing, and Roy Ka-Wei Lee.
\newblock {Math-LLaVA}: Bootstrapping mathematical reasoning for multimodal
  large language models.
\newblock \emph{arXiv preprint arXiv:2406.17294}, 2024.

\bibitem[Sima et~al.(2025)Sima, Renz, Chitta, Chen, Zhang, Xie, Luo, Geiger,
  and Li]{DriveLMSima}
Chonghao Sima, Katrin Renz, Kashyap Chitta, Li Chen, Hanxue Zhang, Chengen Xie,
  Ping Luo, Andreas Geiger, and Hongyang Li.
\newblock {DriveLM}: Driving with graph visual question answering.
\newblock In \emph{ECCV}, 2025.

\bibitem[Singh et~al.(2019)Singh, Natarajan, Shah, Jiang, Chen, Batra, Parikh,
  and Rohrbach]{singh2019towards}
Amanpreet Singh, Vivek Natarajan, Meet Shah, Yu Jiang, Xinlei Chen, Dhruv
  Batra, Devi Parikh, and Marcus Rohrbach.
\newblock Towards {VQA} models that can read.
\newblock In \emph{CVPR}, pages 8317--8326, 2019.

\bibitem[Sun et~al.(2024)Sun, Liu, Bair, and Kolter]{sun2024a}
Mingjie Sun, Zhuang Liu, Anna Bair, and J~Zico Kolter.
\newblock A simple and effective pruning approach for large language models.
\newblock In \emph{ICLR}, 2024.

\bibitem[Team(2024)]{team2024qwen2}
Qwen Team.
\newblock Qwen2 technical report.
\newblock \emph{arXiv preprint arXiv:2407.10671}, 2024.

\bibitem[Wang et~al.(2024)Wang, Bai, Tan, Wang, Fan, Bai, Chen, Liu, Wang, Ge,
  et~al.]{wang2024qwen2}
Peng Wang, Shuai Bai, Sinan Tan, Shijie Wang, Zhihao Fan, Jinze Bai, Keqin
  Chen, Xuejing Liu, Jialin Wang, Wenbin Ge, et~al.
\newblock {Qwen2-VL}: Enhancing vision-language model's perception of the world
  at any resolution.
\newblock \emph{arXiv preprint arXiv:2409.12191}, 2024.

\bibitem[Wang et~al.(2025{\natexlab{a}})Wang, Gao, Gu, Pu, Cui, Wei, Liu, Jing,
  Ye, Shao, et~al.]{wang2025internvl3_5}
Weiyun Wang, Zhangwei Gao, Lixin Gu, Hengjun Pu, Long Cui, Xingguang Wei,
  Zhaoyang Liu, Linglin Jing, Shenglong Ye, Jie Shao, et~al.
\newblock {InternVL3.5}: Advancing open-source multimodal models in
  versatility, reasoning, and efficiency.
\newblock \emph{arXiv preprint arXiv:2508.18265}, 2025{\natexlab{a}}.

\bibitem[Wang et~al.(2025{\natexlab{b}})Wang, Tang, Liu, and
  Xu]{wang2025tracking}
Yubo Wang, Jianting Tang, Chaohu Liu, and Linli Xu.
\newblock Tracking the copyright of large vision-language models through
  parameter learning adversarial images.
\newblock In \emph{ICLR}, 2025{\natexlab{b}}.

\bibitem[Wang et~al.(2023)Wang, Li, Wu, Soon, and Zhang]{wang2023finvis}
Ziao Wang, Yuhang Li, Junda Wu, Jaehyeon Soon, and Xiaofeng Zhang.
\newblock {FinVis-GPT}: A multimodal large language model for financial chart
  analysis.
\newblock \emph{arXiv preprint arXiv:2308.01430}, 2023.

\bibitem[Wu et~al.(2024)Wu, Chen, Pan, Liu, Liu, Dai, Gao, Ma, Wu, Wang, Xie,
  Wu, Hu, Wang, Sun, Li, Piao, Guan, Liu, Xie, You, Dong, Yu, Zhang, Zhao,
  Wang, and Ruan]{wu2024deepseekvl2}
Zhiyu Wu, Xiaokang Chen, Zizheng Pan, Xingchao Liu, Wen Liu, Damai Dai, Huazuo
  Gao, Yiyang Ma, Chengyue Wu, Bingxuan Wang, Zhenda Xie, Yu Wu, Kai Hu, Jiawei
  Wang, Yaofeng Sun, Yukun Li, Yishi Piao, Kang Guan, Aixin Liu, Xin Xie,
  Yuxiang You, Kai Dong, Xingkai Yu, Haowei Zhang, Liang Zhao, Yisong Wang, and
  Chong Ruan.
\newblock {DeepSeek-VL2}: Mixture-of-experts vision-language models for
  advanced multimodal understanding, 2024.

\bibitem[Xu et~al.(2024)Xu, Wang, Ma, Koh, Xiao, and
  Chen]{xu-etal-2024-instructional}
Jiashu Xu, Fei Wang, Mingyu Ma, Pang~Wei Koh, Chaowei Xiao, and Muhao Chen.
\newblock Instructional fingerprinting of large language models.
\newblock In \emph{NAACL}, pages 3277--3306, 2024.

\bibitem[Yadav et~al.(2023)Yadav, Tam, Choshen, Raffel, and
  Bansal]{yadav2023tiesmerging}
Prateek Yadav, Derek Tam, Leshem Choshen, Colin Raffel, and Mohit Bansal.
\newblock {TIES}-merging: Resolving interference when merging models.
\newblock In \emph{NeurIPS}, 2023.

\bibitem[Yang et~al.(2025)Yang, Wu, Shen, Dai, Backes, and
  Zhang]{yang2025challenge}
Ziqing Yang, Yixin Wu, Yun Shen, Wei Dai, Michael Backes, and Yang Zhang.
\newblock The challenge of identifying the origin of black-box large language
  models.
\newblock \emph{arXiv preprint arXiv:2503.04332}, 2025.

\bibitem[Zhang et~al.(2025)Zhang, Liu, Qian, Zhang, Liu, Qiao, and
  Shao]{zhang2025reef}
Jie Zhang, Dongrui Liu, Chen Qian, Linfeng Zhang, Yong Liu, Yu Qiao, and Jing
  Shao.
\newblock {REEF}: Representation encoding fingerprints for large language
  models.
\newblock In \emph{ICLR}, 2025.

\bibitem[Zhao et~al.(2023)Zhao, Pang, Du, Yang, Li, Cheung, and
  Lin]{zhao2023evaluate}
Yunqing Zhao, Tianyu Pang, Chao Du, Xiao Yang, Chongxuan Li, Ngai-Man Cheung,
  and Min Lin.
\newblock On evaluating adversarial robustness of large vision-language models.
\newblock In \emph{NeurIPS}, 2023.

\bibitem[Zheng et~al.(2024)Zheng, Zhang, Zhang, Ye, Luo, Feng, and
  Ma]{zheng2024llamafactory}
Yaowei Zheng, Richong Zhang, Junhao Zhang, Yanhan Ye, Zheyan Luo, Zhangchi
  Feng, and Yongqiang Ma.
\newblock Llamafactory: Unified efficient fine-tuning of 100+ language models.
\newblock In \emph{Proceedings of the 62nd Annual Meeting of the Association
  for Computational Linguistics (Volume 3: System Demonstrations)}, Bangkok,
  Thailand, 2024. Association for Computational Linguistics.

\bibitem[Zhu et~al.(2023)Zhu, Chen, Shen, Li, and Elhoseiny]{zhu2023minigpt}
Deyao Zhu, Jun Chen, Xiaoqian Shen, Xiang Li, and Mohamed Elhoseiny.
\newblock {MiniGPT-4}: Enhancing vision-language understanding with advanced
  large language models.
\newblock \emph{arXiv preprint arXiv:2304.10592}, 2023.

\bibitem[Zhu et~al.(2016)Zhu, Groth, Bernstein, and Fei-Fei]{zhu2016visual7w}
Yuke Zhu, Oliver Groth, Michael Bernstein, and Li Fei-Fei.
\newblock {Visual7W}: Grounded question answering in images.
\newblock In \emph{CVPR}, pages 4995--5004, 2016.

\end{thebibliography}
}

\clearpage
% \clearpage
% \setcounter{page}{1}
% \maketitlesupplementary

% \section{Rationale}
% \label{sec:rationale}
\setcounter{equation}{10}
\setcounter{figure}{5}
\setcounter{table}{7}

% \startcontents[app]
% \section*{Contents}
% \printcontents[app]{l}{1}{}

This supplementary material includes details of original MLLMs, fine-tuning and inference settings, and datasets used in fine-tuning. Furthermore, we provide the inference parameter analysis and utility performance of both original models and fine-tuned models.
The contents are organized as follows:
\begin{itemize}
    \item \S\ref{A1} Details of original models
    \item \S\ref{A2} Details of downstream fine-tuning Datasets
    \item \S\ref{A3} Fine-tuning setting
    \item \S\ref{A4} Inference setting
    % \item \S\ref{B1} Derivations of Eq. (7)
    \item \S\ref{C1} Utility of the fine-tuning models
    \item \S\ref{C2} Impact on inference parameter settings
    \item \S\ref{C3} More multimodal datasets fine-tuned models
    \item \S\ref{C4} Experiments on more MLLMs
    \item \S\ref{C5} Tracking results under input transformations
    \item \S\ref{C6} Ablation study of more parameters
    \item \S\ref{C7} Sensitivity analysis of trigger selection
    \item \S\ref{C8} Robustness of system prompt variations
    \item \S\ref{C9} CLIP-like module stability evidence
\end{itemize}

\appendix
\section{Implementation Details}
\subsection{Details of original models}
\label{A1}
We choose LLaVA-1.5~\cite{liu2024improved} and Qwen2-VL~\cite{wang2024qwen2} as the original MLLMs to obtain derivative MLLMs, and then construct triggers on the original MLLMs.

\noindent\textbf{LLaVA-1.5.} For LLaVA-1.5, we choose LLaVA-1.5-7B, a MLLM of end-to-end training, which consist of a frozen vision encoder CLIP ViT-14L~\cite{radford2021learning}, a visual language connector with two linear layers, and a large language model decoder LLaMA-2 with a total of 32 layers, and the 4096 hidden dimensions. 

\noindent\textbf{Qwen2-VL.} For Qwen2-VL, we choose Qwen2-VL-2B-Instruct, which enables the model to dynamically process images of varying resolutions and also integrates multimodal rotary position embedding, facilitating the effective fusion of positional information. It is an end-to-end unified transformer architecture that integrates vision encoders and language models, and consists of a vision encoder with a resolution of $224\times224$, and a language model Qwen2~\cite{team2024qwen2}.

% \subsection{Experiment Setting}
% 怎么算ASR，inference 的设置，实验的parameter设置, fine-tuning细节
% \subsection{Basic Setting}
% For model parameter optimization, we use the learning rate $\gamma=5e-4$ for LLaVA-1.5. 

% \begin{table}[!t]
%   \centering
%     % \caption*{Table~I: Fine-tuning setting in the experiments.} % 表格标题
%     \caption{Runime(s).} % 表格标题
%   \resizebox{0.50\textwidth}{!}{\begin{tabular}{c|ccc} % 三个居中对齐列 (c c c)
%     \toprule    % 顶部线
%     &RNA&PLA&AGDI  \\
%     \midrule    % 标题与内容分隔线
%      Runtime (s)& 330 & 360&400\\
%     \bottomrule % 底部线
%   \end{tabular}}
%   \label{trainsetting}
% \end{table}

\subsection{Details of downstream fine-tuning datasets}
\label{A2}
% 展示数据：fig+json
In the experiments, we chose five downstream task fine-tuning datasets to simulate various real-world scenarios.
We provide a detailed description of those datasets in the following. Moreover, all train datasets were standardized into the ShareGPT format, specifically designed to simulate natural conversational flows.

\begin{figure*}[!t]
\centering
\includegraphics[width=\linewidth]{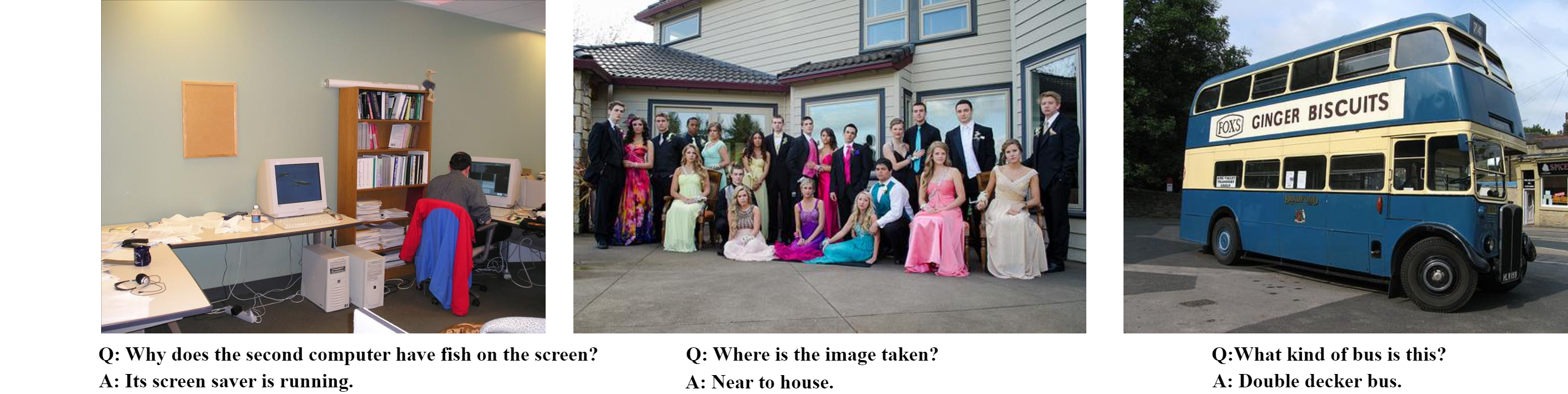}
\caption{Examples of Visual7W dataset.}
    \label{show1}
\end{figure*}

\noindent\textbf{V7W.} Visual7W (V7W)~\cite{zhu2016visual7w} is a dataset designed for comprehensive image content understanding, specifically tailored for VQA tasks. This dataset extends beyond raw images by incorporating region-specific question-answer annotations. It comprises 47,300 COCO-sourced images with 327,929 QA pairs, 1,311,756 human-generated multiple-choice questions, and 561,459 object groundings from 36,579 categories. The questions, structured exclusively as four-option multiple-choice items, are systematically organized around seven interrogative types (What, Where, How, When, Who, Why, Which), as shown in Figure~\ref{show1}.

\begin{table}[!t]
  \centering
    % \caption*{Table~I: Fine-tuning setting in the experiments.} % 表格标题
    \caption{Fine-tuning setting in the experiments.} % 表格标题
  \resizebox{0.50\textwidth}{!}{\begin{tabular}{lll} % 三个居中对齐列 (c c c)
    \toprule    % 顶部线
    Hyperparameter&LoRA fine-tuning setting & Full fine-tuning setting  \\
    \midrule    % 标题与内容分隔线
     Optimizer & AdamW & AdamW\\
     Learning rate& 2e-4&1e-5 \\
      Batch size&8&4 \\
      LoRA rank & 16& / \\
      LoRA alpha& 32& / \\
     Training epochs & 3&3\\
     Gradient accumulation& 1&2\\
      Dtype& bfloat16 &bfloat16 \\
      Lr scheduler& cosine&cosine\\
      Warm-up epoch ratio& 0.03&0.01\\
    \bottomrule % 底部线
  \end{tabular}}
  \label{trainsetting}
\end{table}
\begin{table*}[!t]
  \centering
    % \caption*{Table~II: Utility performance of LoRA fine-tuned LLaVA-1.5 models.} % 表格标题
    \caption{Utility performance of LoRA fine-tuned LLaVA-1.5 models.} % 表格标题
  \begin{tabular}{lccccc} 
    \toprule    % 顶部线
    Datesets& V7W (ACC) &ST-VQA (ACC)& TextVQA (ACC)&PaintingF (BLEU~/~ROUGE)& MathV (ACC) \\
    \midrule    % 标题与内容分隔线
     Before fine-tuning & 0.6\% &21.9\% &15.6\% &4.9\%~/~8.8\% &27.1\% \\
    After fine-tuning& 37.9\%& 65.2\%& 51.3\%& 13.3\%~/~15.9\%&57.5\% \\
    \bottomrule % 底部线
  \end{tabular}
\label{Utility1}
\end{table*}

\begin{table*}[!t]
  \centering
      % \caption*{Table~III: Utility performance of LoRA fine-tuned Qwen2-VL models.} % 表格标题
      \caption{Utility performance of LoRA fine-tuned Qwen2-VL models.} % 表格标题
  \begin{tabular}{lccccc} 
    \toprule    % 顶部线
    Datesets& V7W (ACC) &ST-VQA (ACC)& TextVQA (ACC)&PaintingF (BLEU~/~ROUGE)& MathV (ACC) \\
    \midrule    % 标题与内容分隔线
     Before fine-tuning & 0.8\% &29.8\% & 23.9\%&7.6\%~/~11.9\% &7.0\% \\
    After fine-tuning&37.9\% & 47.7\%&78.7\% &13.4\%~/~16.2\% & 66.6\%\\
    \bottomrule % 底部线
  \end{tabular}
  \label{Utility2}
\end{table*}
\begin{table*}[!t]
  \centering
      % \caption*{Table~IV: Utility performance of full fine-tuned Qwen2-VL models.} % 表格标题
    \caption{ Utility performance of full fine-tuned Qwen2-VL models.} % 表格标题
  \begin{tabular}{lccccc} 
    \toprule    % 顶部线
    Datesets& V7W (ACC) &ST-VQA (ACC)& TextVQA (ACC)&PaintingF (BLEU~/~ROUGE)& MathV (ACC) \\
    \midrule    % 标题与内容分隔线
     Before fine-tuning & 0.8\% &29.8\% & 23.9\%&7.6\%~/~11.9\% &7.0\% \\
    After fine-tuning&36.5\% &88.8\%&78.6\%&13.5\%~/~15.9\%&86.0\%\\
    \bottomrule % 底部线
  \end{tabular}
  \label{Utility3}
\end{table*}

\begin{figure*}[!t]
\centering
\includegraphics[width=0.90\linewidth]{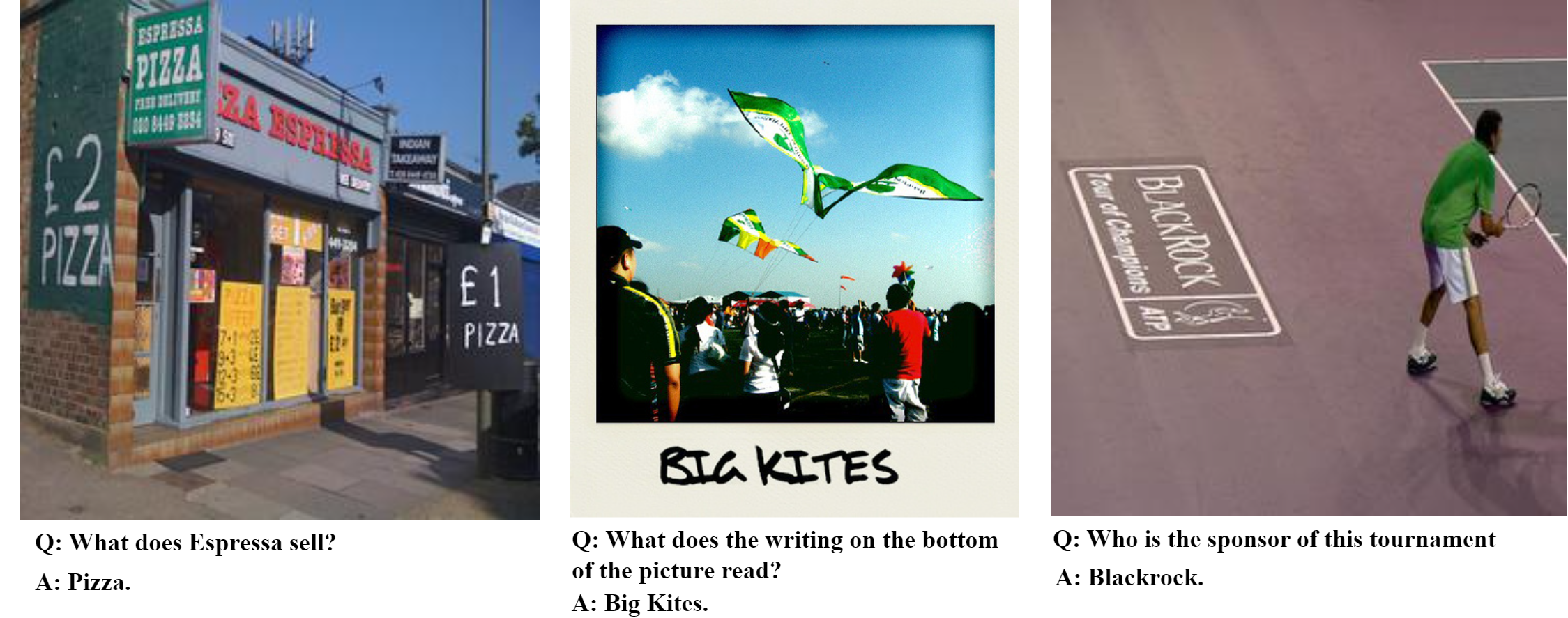}
\caption{Examples of ST-VQA dataset.}
    \label{show2}
\end{figure*}

\noindent\textbf{ST-VQA.} The ST-VQA~\cite{biten2019scene} dataset comprises 23,038 images from six diverse sources (including scene-text benchmarks like COCO-Text and VizWiz, and general vision datasets such as ImageNet and Visual Genome) to mitigate inherent biases and enhance question variety. Each image contains two or more scene text instances, ensuring multiple answer options. The dataset provides 31,791 non-binary, unambiguous question-answer pairs requiring explicit reasoning about textual elements within visual contexts. It is split into training (19,027 images with 26,308 QA pairs) and evaluation subsets for standardized benchmarking, as shown in Figure~\ref{show2}.

\begin{figure*}[!t]
\centering
\includegraphics[width=0.90\linewidth]{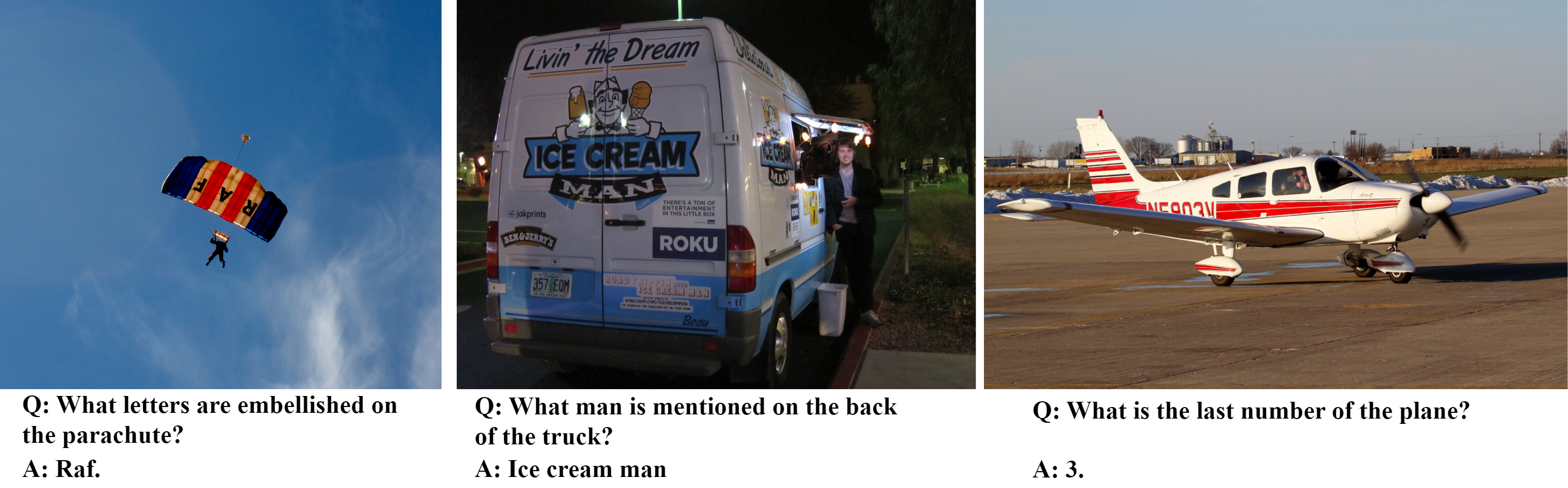}
\caption{Examples of TextVQA dataset.}
    \label{show3}
\end{figure*}

\noindent\textbf{TextVQA.} TextVQA~\cite{singh2019towards} is a standard benchmark for text-based visual reasoning, requiring models to read and reason about scene text within images to answer questions. The dataset comprises 28,408 images and 45,336 questions. It is split into training (21,953 images; 34,602 questions), validation (3,166 images; 5,000 questions), and test sets (3,289 images; 5,734 questions), as shown in Figure~\ref{show3}.

\begin{figure*}[!t]
\centering
\includegraphics[width=0.90\linewidth]{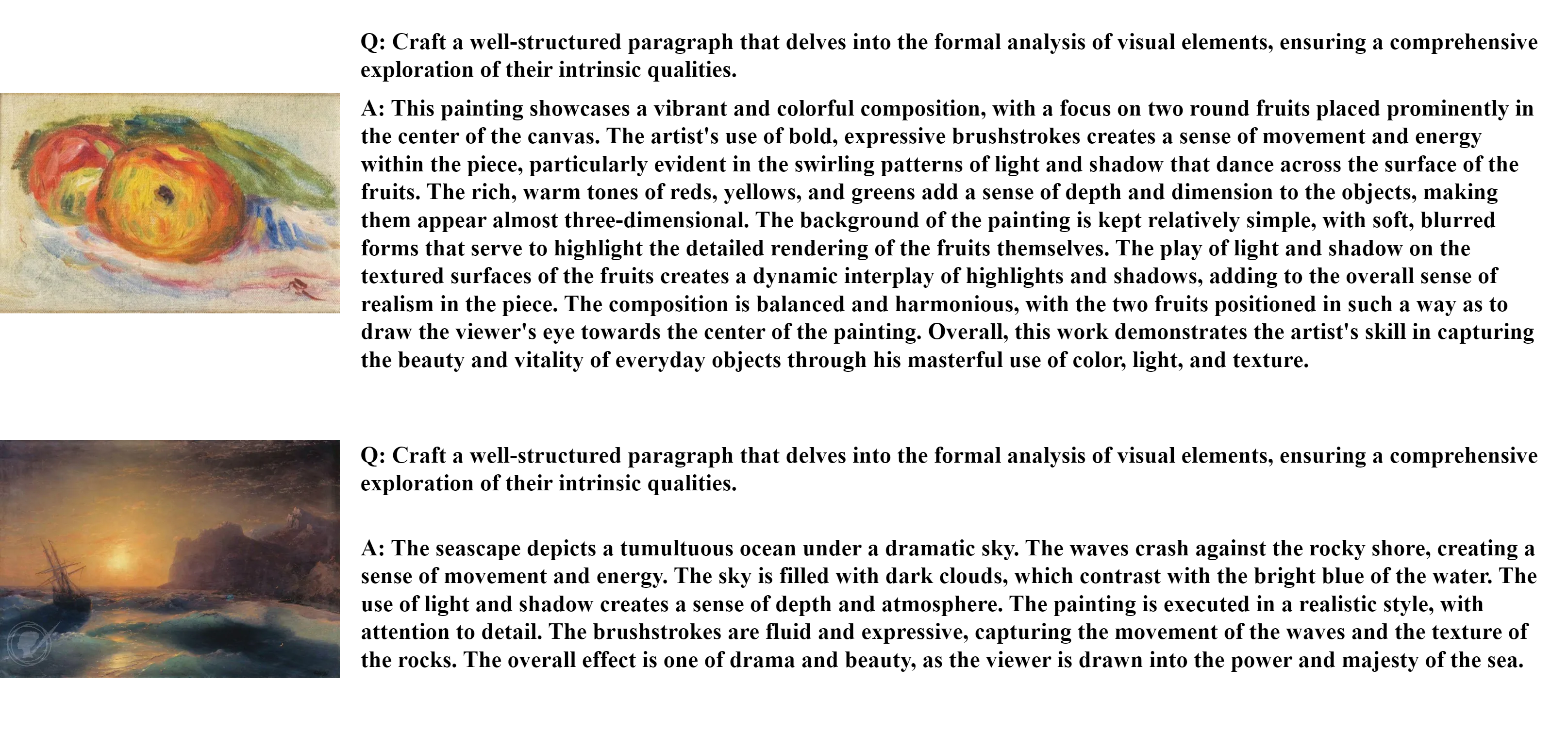}
\caption{Examples of PaintingForm dataset.}
    \label{show4}
\end{figure*}

\noindent\textbf{PaintingForm.} The PaintingForm~\cite{bin2024gallerygpt} dataset comprises 19,000 painting images paired with 50,000 expert analysis paragraphs focused exclusively on visual characteristics of artwork. Designed to advance multimodal AI for deep understanding of artistic elements, as shown in Figure~\ref{show4}.

\begin{figure*}[!t]
\centering
\includegraphics[width=0.90\linewidth]{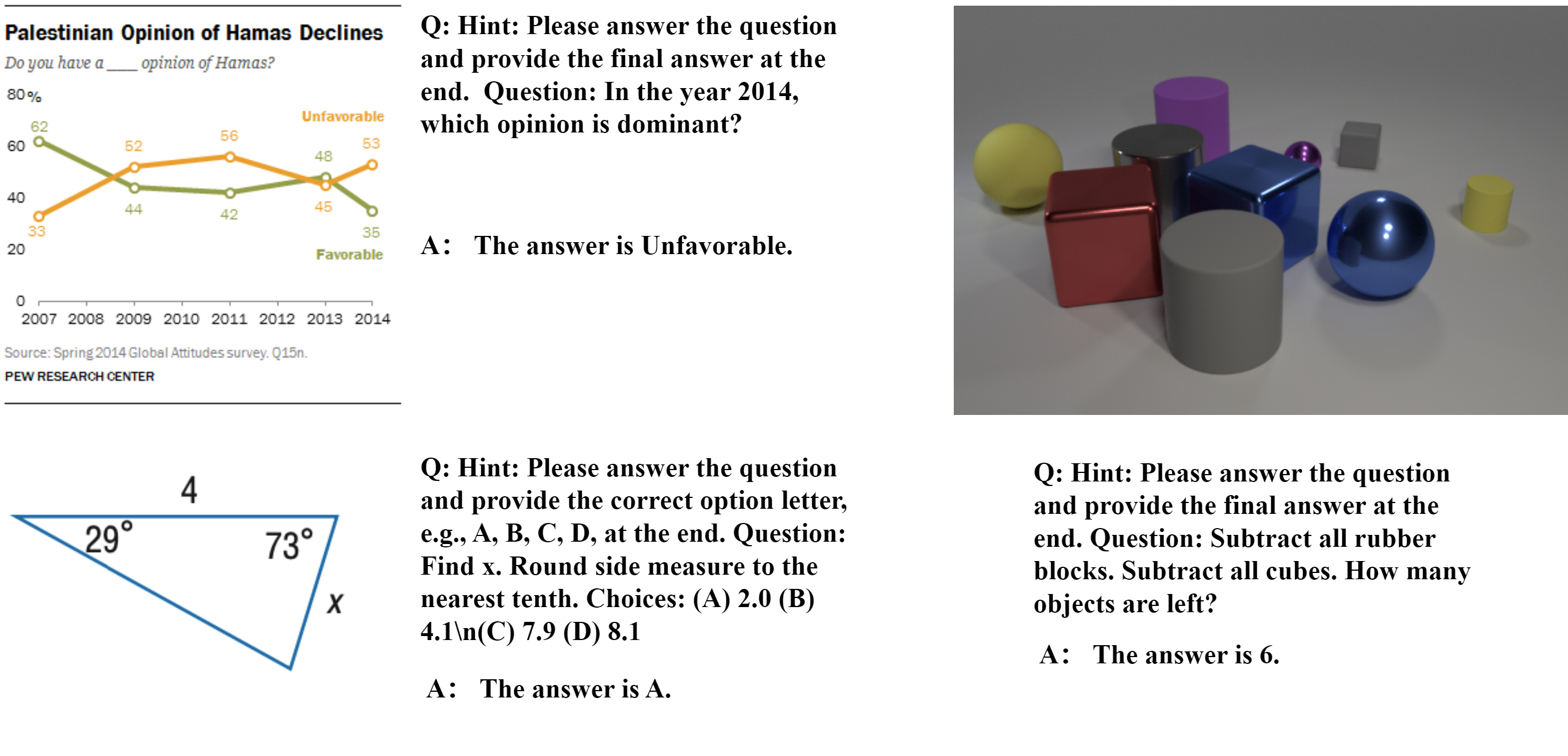}
\caption{Examples of MathV360k dataset.}
    \label{show5}
\end{figure*}

\noindent\textbf{MathV360k.} MathV360K~\cite{shi2024math} is a comprehensive multimodal benchmark synthesized from 24 open-source datasets to advance mathematical visual reasoning. Curated through a rigorous selection process, the dataset originates from 40K high-quality images filtered by visual clarity and cognitive complexity. Each image is enriched with 360K diverse instruction-tuning pairs targeting five high-level reasoning domains: Figure Question Answering (FQA), Geometry Problem Solving (GPS), Math Word Problems (MWP), Textbook Question Answering (TQA), and Visual Question Answering (VQA). This multi-domain architecture addresses critical gaps in existing resources by enhancing image comprehension and mathematical reasoning capabilities, as shown in Figure~\ref{show5}.

\subsection{Fine-tuning setting}
\label{A3}
For both full and LoRA~\cite{hu2022lora} fine-tuning settings, the detailed training configurations are summarized in Table~\ref{trainsetting}. All LoRA fine-tuned models are evaluated in their merged form. For Qwen2-VL, multimodal downstream task fine-tuning is conducted based on the LlamaFactory~\cite{zheng2024llamafactory} project.

\subsection{Inference setting}
\label{A4}
We use ``generate'' function for LLaVA-1.5's inference in all experiments. We set inference parameters such as temperature at 0.5, Top-p at 0.5, num-beams at 1, and max-new-tokens at 128. For Qwen2-VL, we use default inference setting with max-new-tokens at 128.

\begin{figure*}[!t]
    \centering
    \subfloat[\textbf{LLaVA-1.5 (ST-VQA).}\label{llava_stvqa}]{
        \includegraphics[width=0.22\textwidth]{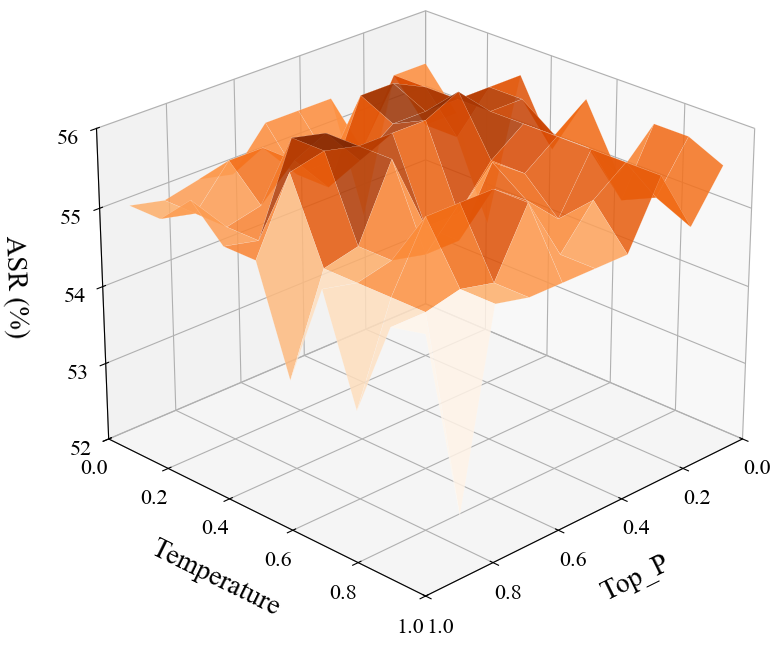}
    }
    \subfloat[\textbf{LLaVA-1.5 (PaintingF).}\label{llava_printingf}]{
        \includegraphics[width=0.22\textwidth]{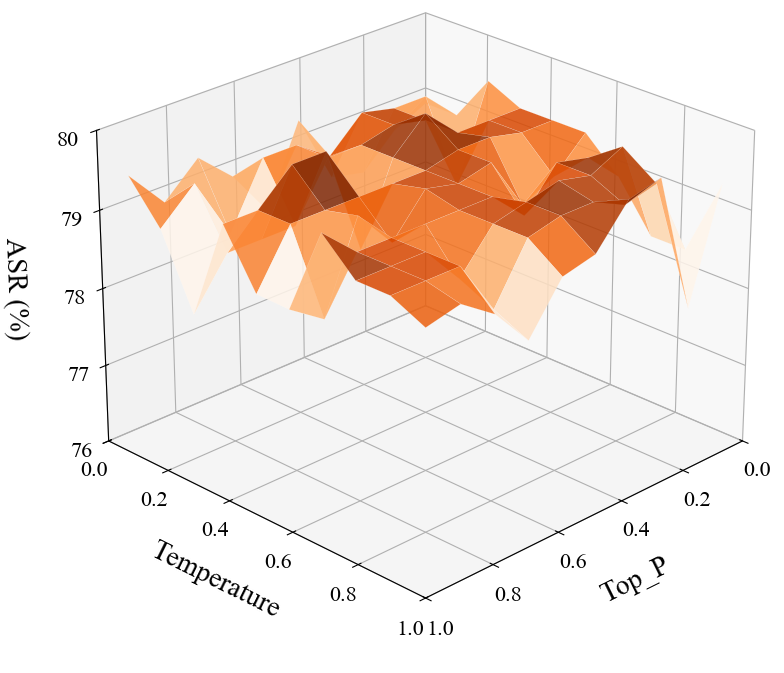}
    }
    \subfloat[\textbf{Qwen2-VL (ST-VQA).}\label{qwenvl_stvqa}]{
        \includegraphics[width=0.22\textwidth]{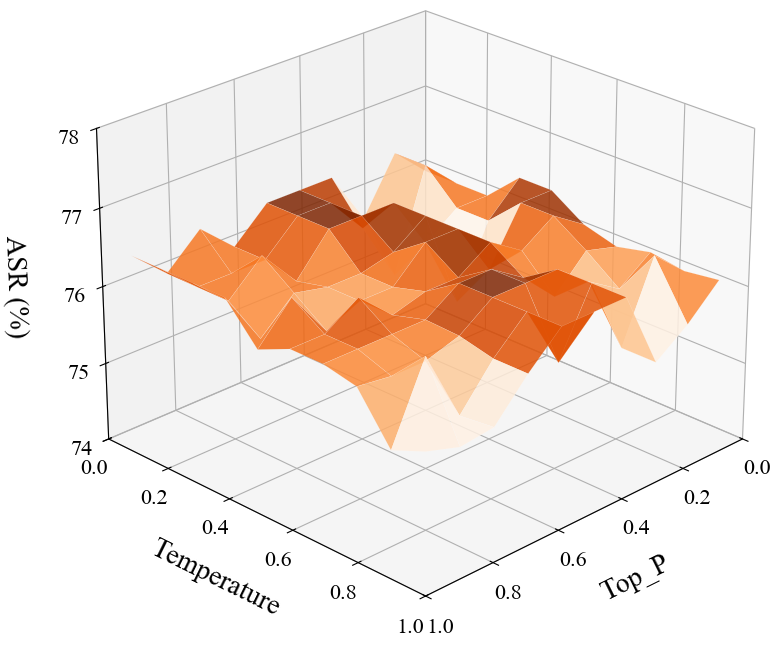}
    }
    \subfloat[\textbf{Qwen2-VL (PaintingF).}\label{qwenvl_printingf}]{
        \includegraphics[width=0.22\textwidth]{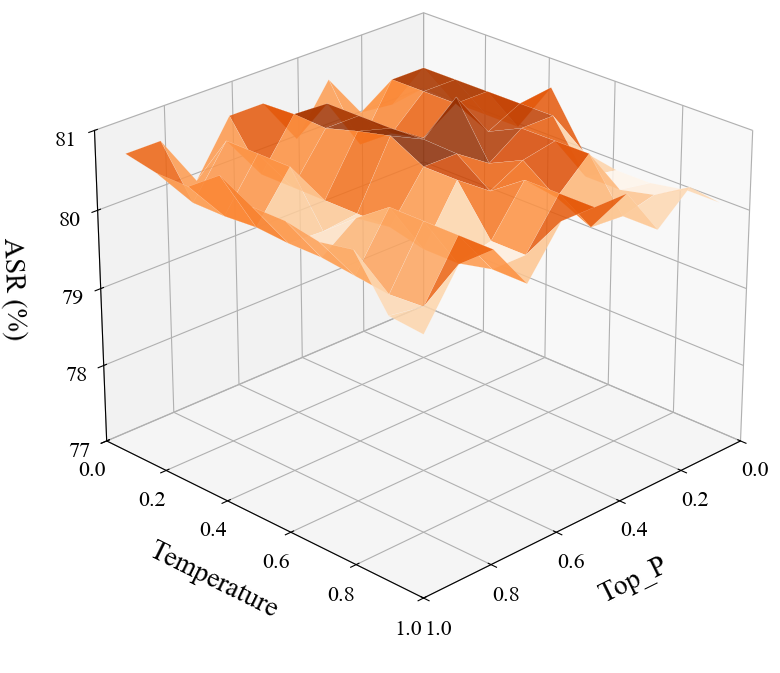}
    }
    \caption{Hyperparameter analysis results of inference settings (Top-p and temperature). (a)(b) ASR on ST-VQA and PaintingF for LLaVA-1.5; (c)(d) ASR for Qwen2-VL.  }
    \label{inferpara}
\end{figure*}
\section{Additional Experiments}

\subsection{Utility of the fine-tuning models}
\label{C1}
We test the utility performance of our two original MLLMs before and after LoRA fine-tuning on downstream task datasets. For full fine-tuning, we report the utility performance of Qwen2-VL before and after fine-tuning on downstream task datasets. As detailed in Tables~\ref{Utility1},~\ref{Utility2}, and~\ref{Utility3} fine-tuning substantially enhanced MLLM performance on target tasks. This indicates that there has been a significant change through fine-tuning the model parameters, demonstrating that fine-tuned models effectively simulate real-world application scenarios. Evaluation was conducted on 5,000 randomly sampled test/validation instances from each dataset. For dataset V7W, ST-VQA, TextVQA, and MathV360k, we computed accuracy (ACC) based on answer correspondence. For the dataset PaintingForm, we evaluated BLEU~\cite{papineni2002bleu} and ROUGE~\cite{lin-2004-rouge} scores.

\subsection{Impact on inference parameter settings}
\label{C2}
In practical model deployment, the performance of trigger tracing may vary due to the randomization effects caused by the configuration of sampling parameters such as temperature and top-P during downstream model inference. We test two fine-tuned variants of LLaVA-1.5 and Qwen2VL across temperature and Top-p values ranging from 0.1 to 1.0 in 0.1 increments. As shown in Figure~\ref{inferpara}, the results
show that the ASR fluctuation stayed within ±1 \%, confirming that AGDI is resilient to inference randomization, such as sampling parameter variation.
\subsection{More multimodal datasets fine-tuned models}
\label{C3}
Tracking the copyright of models fine-tuned for diverse downstream tasks is essential for strong MLLM copyright protection. To further evaluate our method's robustness, we conduct validation across five prevalent task domains: remote sensing, medical, scientific, autonomous driving, and finance. Specifically, we select  RSVQA~\cite{9088993}, DriveLM~\cite{DriveLMSima}, FinVis~\cite{wang2023finvis}, PathVQA~\cite{he2020pathvqa} and SciVerse~\cite{guo2025sciverse}. The details of these five datasets are as follows:
% 模型性能测试之后再说
\begin{table*}[!t]
  \centering
      \caption{ Copyright tracking performance on more multimodal downstream models.}
    \resizebox{0.90\linewidth}{!}{
		\setlength\tabcolsep{5pt}
		\renewcommand\arraystretch{1}
        \begin{tabular}{l||ccccc|ccccc}
            % \thickhline 
            % \toprule 
            \hline
            \rowcolor{mygray}
&\multicolumn{5}{c|}{Qwen2-VL}  &\multicolumn{5}{c}{InternVL3.5}\\
            \cline{2-11}
        % \cmidrule{2-10}
            \rowcolor{mygray} \multirow{-2}{*}{\textbf{Method}}& Pathvqa&DriveVQA & FinVis& RSVQA&SciVerse &Pathvqa&DriveVQA & FinVis& RSVQA&SciVerse  \\ 
            \hline \hline
            % \midrule\midrule
     Ordinary& 41\%& 52\% & 50\% & 55\%& 69\%& 30\%& 27\% &26\%  &36\% & 46\% \\       
    % IF&  &   &   &  &   &  &   &   &    &      \\
    RNA& 39\%& 43\% & 43\% & 46\%& 54\%& 29\%& 19\% & 17\% & 30\%& 35\%  \\
    PLA & 55\%& 76\% & 72\% & 76\%& 86\%& 40\%& 46\% & 52\% &64\% & 73\%  \\
    \hline
\rowcolor{gray!10}
    AGDI& \textbf{58\%}& \textbf{80\%} & \textbf{77\%} & \textbf{79\%}& \textbf{88\%}& \textbf{45\%}& \textbf{50\%} & \textbf{57\%} & \textbf{70\%}& \textbf{76\%} \\
            \hline
            % \bottomrule
        \end{tabular}}  
\label{moredata_track}
\end{table*}
\begin{itemize}
    \item \textbf{RSVQA} is the first remote sensing VQA dataset constructed by automatically extracting information from OpenStreetMap. It comprises two versions based on low-resolution Sentinel-2 satellite imagery and high-resolution aerial imagery, respectively, and covers five types of question-answer pairs: counting, existence, area estimation, comparison, and urban-rural classification.
\item \textbf{DriveLM} is a graph-structured VQA dataset for autonomous driving, available in both real-world and simulated versions. By integrating semi-regularized with fully automated annotation, it achieves superior scale, coverage, and logical complexity over existing benchmarks, offering a generalizable platform for training and evaluating vision-language models in autonomous driving domain.
\item \textbf{FinVis} presents the first two-stage multimodal instruction dataset designed for financial chart analysis. Featuring a pretraining stage for vision-language alignment on historical charts and an innovative instruction-tuning stage that incorporates future data for forecasting, the dataset is structured as {image, instruction, answer} triplets. It provides comprehensive support for professional tasks including chart description, financial question answering, and trend prediction. 
\item \textbf{PathVQA} is a medical visual question answering dataset derived via a semi-automated pipeline from textbooks and digital libraries. It is designed to simulate the American Board of Pathology examinations, with open-ended clinical questions constituting 50.2\% of its content.
\item \textbf{SciVerse} introduces a multimodal scientific assessment dataset spanning physics, chemistry, and biology. It analyzes LLM capabilities in knowledge, vision, and reasoning by varying the knowledge and visual complexity of the problems. This is paired with a novel scientific CoT evaluation strategy to progressively pinpoint knowledge and logic errors, providing deep diagnostic insights into models' problem-solving gaps.
\end{itemize}

In the fine-tuning setup, we employ the full training set for PathVQA, whereas a 30k-sample subset of the training set is used for each remaining dataset. As shown in Table~\ref{moredata_track}, we report the ASR of LoRA fine-tuned variants of Qwen2-VL-2B-Instruct and InternVL3.5-2B-HF. The results demonstrate that AGDI achieves superior copyright tracing performance compared with the baselines across a wide range of multimodal and fine tuning scenarios.

\subsection{Experiments on more MLLMs}
\label{C4}
We also evaluate the copyright tracking performance on the advanced MLLM such as InternVL3.5~\cite{wang2025internvl3_5}. InternVL3.5 is an open-source multimodal model series featuring a "ViT-MLP-LLM" architecture that significantly enhances reasoning capabilities via cascade reinforcement learning and innovatively introduces a Visual Resolution Router (ViR) for dynamic visual token compression together with Decoupled Vision-Language Deployment. We report the ASR on the InternVL3.5-HF 2B and 8B parameter scale models. As shown in Tables~\ref{internVL8B} and~\ref{intern3.5VL_track}, consistent performance confirms that our method generalizes effectively across different architectures and increasing parameter scales. These results demonstrate that the effectiveness of our method is not limited by model size or architecture, exhibiting strong generalization to fine-tuned models. 
\begin{table}[h]
\caption{Copyright tracking performance on InternVL3.5 8B fine-tuned models.}\label{internVL8B}

\centering
\setlength\tabcolsep{5pt}
\resizebox{0.48\textwidth}{!}{\begin{tabular}{l|l||ccccc}
\hline
 \rowcolor{mygray} \textbf{MLLM}& \textbf{Method}& V7W &ST-VQA& TextVQA&PaintingF& MathV \\
 % \cmidrule{2-7}
% &\multirow{2}{*}{ASR}
% &CLIP score $\uparrow$&CLIP score$\uparrow$  &CLIP score$\uparrow$  \\ 
% \midrule
\hline\hline
% \multirow{3}{*}{InternVL3.5-2B}
% &RNA&23\% &40\%&30\%&10\%&20\%\\
% &PLA&39\% &53\%&45\%&23\%&34\% \\
% &\rowcolor{gray!10}AGDI&\textbf{42\%} &\textbf{55\%}&\textbf{47\%}&\textbf{30\%}&\textbf{41\%} \\
% \cline{1-7}
\multirow{3}{*}{InternVL3.5-8B}
&Ordinary&14\% &21\%&15\%&9\%&16\%\\
&RNA&16\% &22\%&15\%&11\%&20\%\\
&PLA&44\% &55\%&46\%&31\%&49\% \\
\rowcolor{gray!10}&AGDI&\textbf{58\%} &\textbf{63\%}&\textbf{62\%}&\textbf{45\%}&\textbf{58\%} \\
% \bottomrule
\hline
\end{tabular}}

\end{table}

\begin{table*}[!t]
  \centering
      \caption{Copyright tracking performance on InternVL3.5 2B fine-tuned models.}
    \resizebox{0.90\linewidth}{!}{
		\setlength\tabcolsep{5pt}
		\renewcommand\arraystretch{1}
        \begin{tabular}{l||ccccc|ccccc}
            % \thickhline 
            % \toprule 
            \hline
            \rowcolor{mygray}
&\multicolumn{5}{c|}{LoRA Fine-tuning}  &\multicolumn{5}{c}{Full Fine-tuning}\\
            \cline{2-11}
        % \cmidrule{2-10}
            \rowcolor{mygray} \multirow{-2}{*}{\textbf{Method}}& V7W &ST-VQA& TextVQA&PaintingF& MathV &V7W &ST-VQA& TextVQA&PaintingF& MathV  \\ 
            \hline \hline
            % \midrule\midrule
     Ordinary& 25\%& 38\% & 32\% &12\% & 23\%& 35\%& 38\% &30\%  & 21\%& 30\% \\       
    % IF&  &   &   &  &   &  &   &   &    &      \\
    RNA& 23\%& 40\% & 30\% &10\% & 20\%& 33\%& 40\% & 31\% &18\% &  26\% \\
    PLA &39\% & 53\% & 45\% & 23\%& 34\%& 46\% & 53\% &42\% &38\% &45\%  \\
    \hline
\rowcolor{gray!10}
    AGDI& \textbf{42\%}& \textbf{55\%} & \textbf{47\%} &\textbf{ 30\%}& \textbf{41\%}& \textbf{49\%} & \textbf{56\%} & \textbf{44\%} &\textbf{49\%} & \textbf{51\%} \\
            \hline
            % \bottomrule
        \end{tabular}}  
\label{intern3.5VL_track}
\end{table*}
\subsection{Tracking results under input transformations.}\label{C5}
We report the copyright tracing results of our method under several common input level perturbations, including JPEG compression, Gaussian noise, and image resizing, to evaluate the robustness of the generated trigger images. The maximum magnitude of the Gaussian noise is set to 5, and the resized image resolution is fixed at 256. In Table~\ref{input-level}, results on LLaVA-1.5 variants demonstrate that our method exhibits robustness against input transformations.

\begin{table}[h]
\caption{Copyright tracking under input transformations.}
\centering
\setlength\tabcolsep{5pt}
\renewcommand\arraystretch{1}
\resizebox{0.48\textwidth}{!}{\begin{tabular}{l||ccccc}
\hline
 \rowcolor{mygray} & V7W &ST-VQA& TextVQA&PaintingF& MathV  \\
 % \cmidrule{2-7}
% &\multirow{2}{*}{ASR}
% &CLIP score $\uparrow$&CLIP score$\uparrow$  &CLIP score$\uparrow$  \\ 
% \midrule
\hline\hline

Original& 64\%&56\%&36\%&79\%&30\%\\
Resizing& 42\%&40\%&26\%&53\%&19\% \\
Gaussian& 60\%&53\%&31\%&73\%&27\% \\
JPEG& 40\%&39\%&24\%&42\%& 16\%\\
% \bottomrule
\hline
\end{tabular}}

\label{input-level}
\end{table}

\subsection{Ablation study of more parameters}
\label{C6}
\begin{figure*}[!t]
    \centering
    \subfloat[\textbf{Parameter $\lambda$}\label{lambda}]{
        \includegraphics[width=0.40\textwidth]{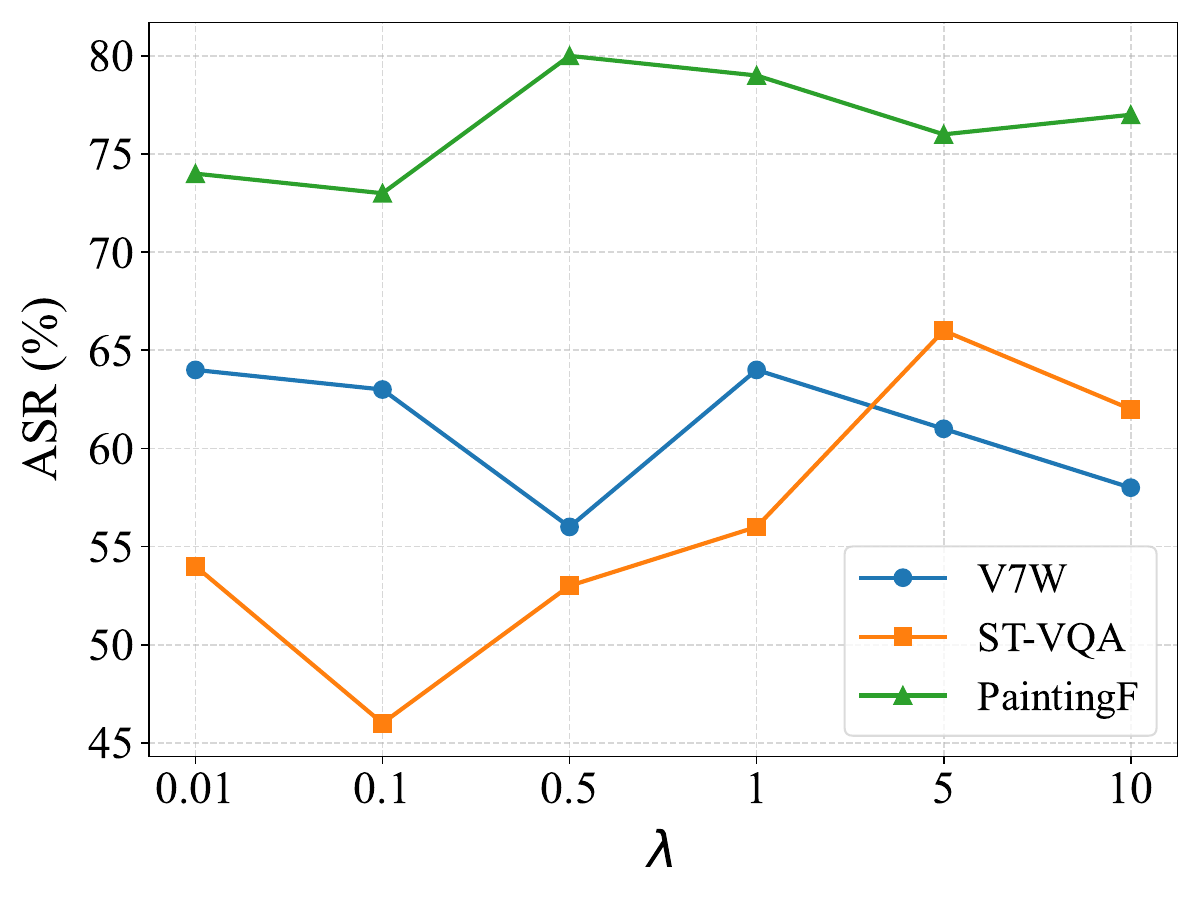}
    }
    ~
    \subfloat[\textbf{Fine-tuning epochs}\label{epochs}]{
        \includegraphics[width=0.40\textwidth]{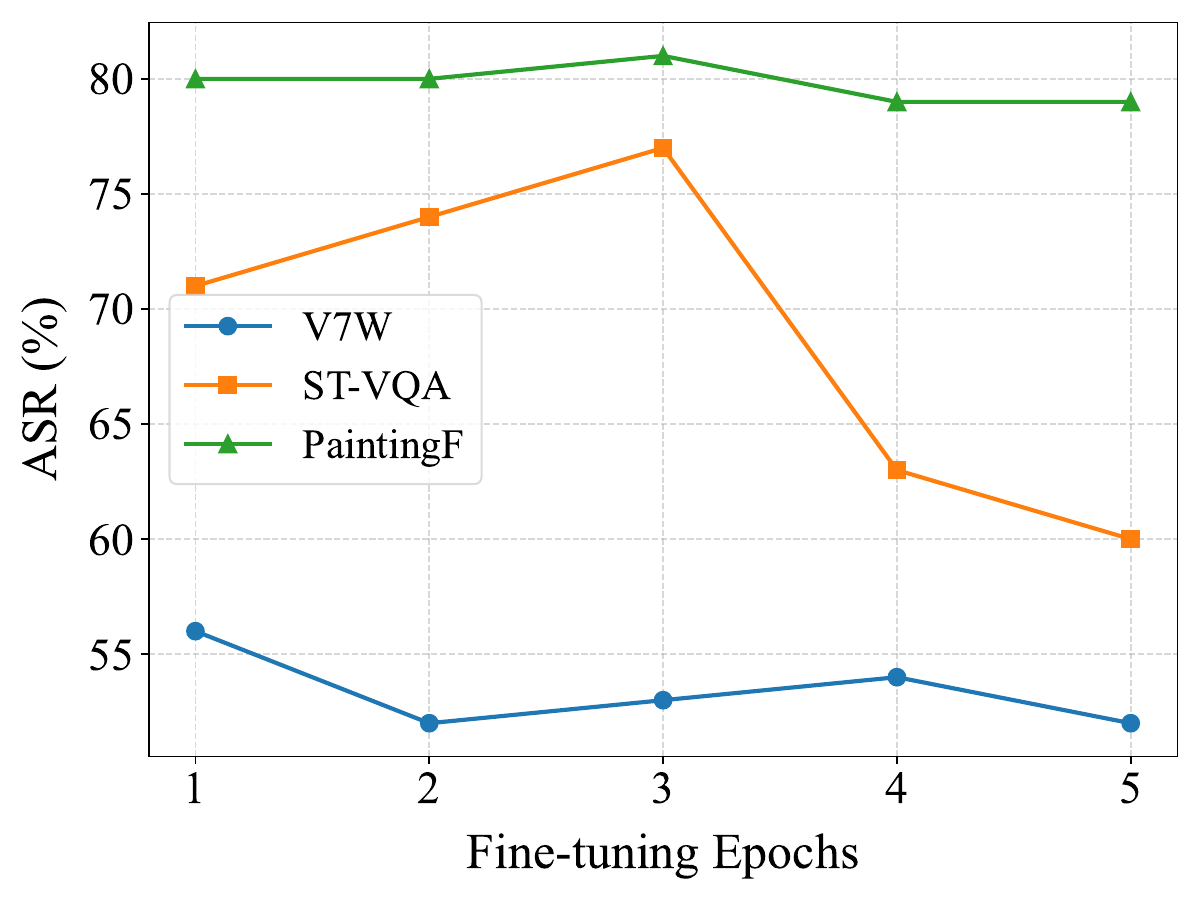}
    }
    \caption{Ablation study for $\lambda$ and fine-tuning epochs: (a) The impact of loss function parameter $\lambda$ on tracking performance; (b) The impact of downstream fine-tuning epochs on tracking performance. }
    \label{parameterexp}
\end{figure*}

We add the ablation study for parameter $\lambda$ and fine-tuning epochs in Figure~\ref{parameterexp}. The results in Figure~\ref{lambda} show consistent and stable performance across various $\lambda$ values, demonstrating the robustness of our method to this hyperparameter. The results in Figure~\ref{epochs} shows that ASR stabilizes with more finetuning, proving increased scale cannot bypass our protection. Our setup aligns with practical scenarios relying on lightweight methods. Furthermore, given the trade-off between cost and utility, aggressive operations like full retraining are impractical because they degrade the model's performance and commercial value.

\subsection{Sensitivity analysis of trigger selection}
\label{C7}
We provide a sensitivity analysis of trigger selection for copyright tracking performance. Table~\ref{per-pair} shows that tracking results depend more on the specific model than the trigger pairs. Furthermore, we use five diverse QA pairs and 200 random images per pair to prevent bias and ensure consistent results on the models.

\begin{table}[h]
\caption{ASR on 5 different QA pairs.}
\centering
% \footnotesize % 或者使用 \footnotesize 进一步缩小
% \small
\setlength\tabcolsep{8pt} % 稍微增加列间距，避免左右太空
\renewcommand\arraystretch{1.1} % 稍微增加行高，视觉更舒适
\begin{tabular}{l||ccccc}
\hline
\rowcolor{mygray} & QA1 & QA2 & QA3 & QA4 & QA5 \\
\hline\hline
V7W & 63\% & 75\% & 70\% & 68\% & 37\% \\
ST-VQA & 58\% & 62\% & 44\% & 61\% & 48\% \\
PaintingF & 81\% & 86\% & 69\% & 93\% & 62\% \\
\hline
\end{tabular}
\label{per-pair}
\end{table}

\subsection{Robustness of system prompt variations}
\label{C8}
In real-world scenarios, malicious users or downstream developers usually modify the system prompt. We test the ASR on the fine-tuned models under system prompt variations. We design two system prompt as follows.
\begin{itemize}
    \item As a clinical psychologist, use an empathetic tone and prioritize asking questions to guide emotional expression before offering advice.
    \item  As a technical interviewer from a top tech firm, evaluate the candidate's programming basics and provide feedback after each answer.
\end{itemize}
Table~\ref{sys-prompt} shows consistent ASR across different models before and after system prompt modifications. These results confirm system prompt robustness of AGDI. 
\begin{table}[h]
\caption{System prompt experiments on LLaVA-1.5 LoRA variants.}
\centering
\setlength\tabcolsep{5pt}
\renewcommand\arraystretch{1}
\resizebox{0.48\textwidth}{!}{\begin{tabular}{l||ccccc}
\hline
 \rowcolor{mygray} & V7W &ST-VQA& TextVQA&PaintingF& MathV  \\
 % \cmidrule{2-7}
% &\multirow{2}{*}{ASR}
% &CLIP score $\uparrow$&CLIP score$\uparrow$  &CLIP score$\uparrow$  \\ 
% \midrule
\hline\hline

Original&64\% &56\%&36\%&79\%&30\%\\
Sys prompt1&61\% &53\%&32\%&73\%&26\% \\
Sys prompt2&63\% &54\%&32\%&74\%&25\% \\
% \bottomrule
\hline
\end{tabular}}
\label{sys-prompt}
\end{table}
\begin{table}[h]
\caption{Average similarity drift(\%) on the fine-tuned models.}
\centering
\setlength\tabcolsep{5pt}
\renewcommand\arraystretch{1}
\resizebox{0.48\textwidth}{!}{\begin{tabular}{l||ccccc}
\hline
 \rowcolor{mygray} Similarity drift& Pair1&Pair2&Pair3&Pair4&Pair5  \\
 % \cmidrule{2-7}
% &\multirow{2}{*}{ASR}
% &CLIP score $\uparrow$&CLIP score$\uparrow$  &CLIP score$\uparrow$  \\ 
% \midrule
\hline\hline
Base$\rightarrow$ V7W& 4.4\%&9.3\%&1.0\%&2.1\%&6.1\%\\
Base$\rightarrow$ MathV&6.9\% &7.1\%&3.3\%&0.5\%&1.2\% \\
% \bottomrule
\hline
\end{tabular}}
\label{sim-drift}
\end{table}

\subsection{CLIP-like module stability evidence}
\label{C9}
We measure the similarity drift between 200 triggers and target texts across various fine-tuned models. As shown in Table~\ref{sim-drift}, the minimal cosine similarity drift strongly supports the CLIP-like semantic stability assumption, confirming that AGDI’s gains stem from exploiting intrinsic model properties.

% \subsection{Computation analysis}

% \subsection{}
% % Qwen2VL的消融实验

% \subsection{}
% % Impact of lambda. 

% \subsection{}
% % Impact of beta.
\end{document}